\documentclass[11 pt, oneside]{article}

\usepackage[pdftex]{graphicx}
 \pdfoutput=1

\usepackage{url}
\usepackage{hyperref}

\usepackage{geometry}
\usepackage{graphicx}
\usepackage{setspace}
\usepackage{natbib}

\usepackage{booktabs}
\usepackage {amsbsy}
\usepackage {amssymb}
\usepackage {amsmath}
\usepackage{multirow}
\usepackage{algorithmic}
\usepackage{algorithm}
\usepackage{enumerate}
\usepackage[affil-sl,auth-sc]{authblk}
\usepackage{color}

\usepackage{natbib}
 \bibpunct{(}{)}{;}{a}{,}{,}
\begin{document}

\title{Empirical Similarity for Absent Data Generation in Imbalanced Classification}

\author{Arash Pourhabib  \\
School of Industrial Engineering and Management\\ Oklahoma State University, Stillwater, OK 74078  \\
email: \texttt{arash.pourhabib@okstate.edu}} 

\maketitle

\begin{center}
\textbf{Abstract}
\end{center}

 When the training data in a two-class classification problem is overwhelmed by one class, most classification techniques fail to correctly identify the data points belonging to the underrepresented class. This paper proposes Similarity-based Imbalanced Classification (SBIC) that simultaneously optimizes the weights of the empirical similarity  function and identifies the locations of absent data points, i.e. unobserved data points from the minority class. Similar to cost-sensitive approaches, SBIC operates on an algorithmic level to handle imbalanced structures and similar to synthetic data generation approaches, it utilizes the properties of  unobserved data points. The results of applying the proposed method to imbalanced  datasets suggests that SBIC is comparable to, and in some cases outperforms, other commonly used classification  techniques for imbalanced datasets.


\vspace*{.3in}

\noindent\textsc{Keywords}: {Empirical similarity, imbalanced classification, synthetic data generation.}

\newpage
\section {Introduction }\label{sec_intro}

Classification is the task of identifying  class labels for features belonging to a specific set of classes. A successful classification algorithm depends upon having a sufficient number of samples for each class. For instance, in a two-class classification, if the training dataset samples from one class significantly outnumber those from the second class, the classification algorithm may fail to correctly identify  the data points belonging to the minority  class, i.e.  the class that has very few representatives in the training dataset. This type of classification is termed imbalanced classification, to reflect the imbalanced nature of the training dataset~\citep{He:09}. In practice, identifying the points belonging to the minority class is more important~\citep{byon:10}, e.g., for warranty applications, when the cost associated with not predicting a warranty claim in advance is much higher than incorrectly predicting the possible occurrence of a claim. 

Many attempts to address imbalanced classification rely on synthetic oversampling, which artificially creates extra data points from the minority class. An important limitation is that these techniques consider data generation and classification as two  independent  tasks.  In other words, they only alter the  dataset, not the algorithm.  Overcoming the limitation requires generating the synthetic data (maybe implicitly) using a mechanism that accounts for the imbalanced nature of the original data. This type of implicit synthetic data generation is termed absent data generation (ADG)~\citep{pourhabib2015absent}. ADG attempts to identify the locations of the synthetic data points which improve the algorithm’s performance but without necessarily generating the points themselves.

ADG, however, is restricted to the specific formulation of kernel Fisher discriminant analysis~\citep{Mika:99}.  Thus, if absent data can only be utilized by using a discriminant analysis  as a base classifier, the inclusion of ADG may only marginally improve  the classification rate. In this paper we show that ADG can be extended beyond the specific formulation of kernel Fisher discriminant analysis which can help achieve better classification for a larger class of datasets.

Similarity-based approaches define an empirical similarity function which assesses the degree of resemblance between different inputs\citep{gilboa2011similarity,gilboa2006empirical}. The method proposed in this paper uses the concept of similarity  to locate absent data points. For every new input, i.e., the test data point, our algorithm uses the weighted average of the training data points, where the weights are determined based on the empirical similarity function.

We show how an empirical similarity function can identify the location of synthetic data having a high degree of similarity  to the existing minority  data points.  To make the synthetic data useful, we impose constraints so that the new data points are close to the boundary of the two classes. The proposed algorithm simultaneously learns the location of absent data and the parameters of the similarity  function. As such, it does not need to generate synthetic data, but instead utilizes the points to obtain a better classifier for imbalanced datasets. This paper makes two contributions to the literature on imbalanced classification. First, it shows that absent data can be generated using a similarity function. Second, the application of the proposed algorithm to the real dataset demonstrates that it 
is competitive to the state-of-the-art methods in imbalanced classification. 

The remainder of this paper is organized  as follows. Section~\ref{sec_lit} briefly reviews the relevant literature. Section~\ref{sec_SBIC} reviews the concept of empirical similarity, formally defines the problem, and presents our approach for imbalanced classification. Section~\ref{sec_numerical} compares the performance of the proposed algorithm with two commonly used techniques using real datasets. Section~\ref{sec_concl} concludes and offers suggestions for future research.

\section{Related Work}\label{sec_lit}
We begin by assuming that the number of data points for one class, the minority, is either absolutely small, i.e., we have very few samples from that class, or is smaller relative to the second class, the majority.  We call the former case absolute imbalance and the latter case relative imbalance, and briefly review the two streams of literature.

An extreme idea for handling imbalanced datasets is to completely remove the minority points in the training stage based on the understanding that a small number of samples from the minority class may not be useful for identifying the boundary between the two classes. Instead,  the focus is identifying the tightest bound for the majority class~\citep{Park:10}. This class of approaches, known as novelty detection, can be useful when there are very few data points from the minority  class. Empirical results suggest, however, that for most imbalanced datasets, especially if the dataset is relatively imbalanced, novelty detection methods are inferior compared with methods that utilize the minority samples~\citep{pourhabib2015absent}.

Resampling methods utilize the minority samples, e.g., bootstrapping~\citep{efron1982jackknife} partially  mitigates the effect of a low minority  presence.  In~\citep{galar2012review} an ensemble classifier takes advantage of having several datasets for learning.  While these approaches can be effective for some specific data structures~\citep{byon:10,chen:05}, resampling the information embedded in the location of a minority data point several times may cause the classifier to overemphasize the region, thus introducing significant bias. In addition, resampling techniques do not allow for ``exploring'' regions which do not have any actual minority points.

The drawbacks of resampling have motivated the concept of synthetic oversampling~\citep{Chawla:02,Han:05,chen2010ramoboost,barua2014mwmote}, which generates extra data points based on the existing data  in order to create an augmented, and less imbalanced, dataset.  Synthetic oversampling methods differ based on the mechanisms they employ for data generation. For example, SMOTE~\citep{Chawla:02} uses linear interpolation between existing minority data points to generate new samples, whereas Borderline-SMOTE~\citep{Han:05} utilizes both minority and majority points to create new samples close to the boundaries of the two classes. Synthetic oversampling can also be combined with undersampling for improved performance~\citep{ramentol2012smote}. 

Another stream of literature focuses on cost-sensitive methods, which  modify the algorithm, rather than the dataset, by assigning imbalanced costs to mis-classification~\citep{Elkan:01,Ting:02,masnadi2010risk}. For example, in the cost sensitive support vector machine, the constraints in the optimization problem are such that the cost associated with labeling a minority data point in training as majority is much higher than that for a majority. Some methods combine cost-sensitive with over/undersampling~\citep{zhou2006training}, or employ cost-sensitive boosting~\citep{sun2007cost}. 

We note that cost-sensitive methods alter the algorithm, whereas synthetic oversampling methods change the dataset; however, we can have a synthetic data generation mechanism that works on an algorithm-level: if we generate data such that the data generation mechanism is embedded in the algorithm, as opposed to have an independent data generator and a classifier, we can obtain an algorithm that generates synthetic data, or \textit{absent data} in this context, to better identify the boundary~\citep{pourhabib2015absent}. 

Section~\ref{sec_SBIC} presents how we employ the idea of absent data generation embedded in a similarity-based algorithm. Our contribution is to demonstrate that absent data generation is not confined to  the formulation of kernel Fisher discriminant analysis (KFDA)~\citep{Mika:99}. To wit, the algorithm in~\citep{pourhabib2015absent} utilizes absent data when the base classifier is KFDA. So, if for a specific data structure KFDA performs very poorly, generating absent data may only marginally improve the classification performance. By extending the application of the idea of absent data beyond KFDA, we demonstrate the versatility of absent data generation for imbalanced classification on a larger class of datasets.
\section{Similarity-based Imbalanced Classification}\label{sec_SBIC}

Supervised learning refers to identifying a behavior in a system, manifested through a function, by empirical means. Supervised learning methods endeavor to generalize based on the information embedded in data, i.e., they employ inductive reasoning~\citep{de1998machine} and then they establish rules which can be utilized to characterize the system, and predict its behavior. At the heart of this generalization is the notion of \textit{empirical similarity} (hereafter, similarity): examine the historical data for similarities between current and previous settings and use the similarities to predict the system’s behavior~\citep{gilboa2011similarity}. 

Assume a training dataset  $D=\{(\mathbf{x}_1,y_1),(\mathbf{x}_2,y_2),\dots,(\mathbf{x}_n,y_n)\}$ where $\mathbf x_i\in\mathbb{R}^p$ is an input, or the system's setting as discussed above, and $y_i\in\mathbb{R}$ is the system's response, or behavior, for $i=1,\dots,n$. That is, we have $n$ input-output observations based on which one can make a generalization about the system's behavior. We know there exists a function $f$ such that $y_i=f(\mathbf x_i)$, for $i=1,\ldots,n$. The objective is to determine this function based on the information in $D$ in order to predict the system's behavior at an unseen location $\mathbf x_t$, i.e. $y_t=f(\mathbf x_t)$. Denote this predicted value as $\widehat{y}_t$. Assume  a function $S:\mathbb{R}^{p}\times\mathbb{R}^{p}\mapsto\mathbb{R}$, where $S(\mathbf x,\mathbf x')$ measures the similarity of $\mathbf x$ to $\mathbf x'$. A straightforward application of similarity-based reasoning suggests
\begin{eqnarray}
\widehat{y}_t=\frac{\sum_{i=1}^n S(\mathbf x_i,\mathbf x_t)y_{t}}{\sum_{i=1}^n S(\mathbf x_i,\mathbf x_t)},\label{eq:gen}
\end{eqnarray}
which means the predicted value $\widehat{y}_t$ is a weighted average of the observations in $D$ where the weights are the similarity measured by $ S(\mathbf x_i,\mathbf x_t)$ for $i=1,\ldots,n$. The idea of similarity-based prediction is related to some other statistical predictive models such as kernel regression, Bayesian updating, and interpolation (see~\citep{gilboa2006empirical} for a discussion) . The expression in equation~\eqref{eq:gen} presents the prediction approach intuitively. Note that equation~\eqref{eq:gen}, which is in a very general form, needs be to tailored to fit imbalanced classification, the focus of this paper. The following sub-sections discuss a form for the similarity  function $S(\mathbf x,\mathbf x')$ and a synthetic data generation for similarity-based classification.

\subsection{Similarity-based classification}
We focus on a two-class classification where the function value $y=f(\mathbf x)$ has only two values, or labels, $0$ or $1$. Recall that  $D=\{(\mathbf{x}_1,y_1),(\mathbf{x}_2,y_2),\dots,(\mathbf{x}_n,y_n)\}\subset\mathcal{X}\times\mathcal{Y}$ denotes the training dataset, where $\mathcal{X}$ is the input domain, and $\mathcal{Y}$ is the output domain. In a two-class classification, we can partition the set $D$ into $D^-$ and $D^+$ such that $D^-\subset\mathbb{R}^{p+1}$ contains only the data points labeled $0$, and $D^+\subset\mathbb{R}^{p+1}$ contains only the data points labeled $1$. Obviously, $D^-\cup D^+=D$ and $D^-\cap D^+=\phi$. Without loss of generality, assume the data points are indexed so that the first $n^-$ data points belong to $D^-$ and the remaining $n^{+}=n-n^{-}$ belong to $D^+$. Specifically, $D^-=\{(\mathbf{x}_1,y_1),(\mathbf{x}_2,y_2),\dots,(\mathbf{x}_{n^-},y_{n^-})\}$ where $y_i=0$ for $i=1,\ldots,n^-$, and $D^+=\{(\mathbf{x}_{n^-+1},y_{n^-+1}),(\mathbf{x}_{n^-+2},y_{n^-+2}),\dots,(\mathbf{x}_n,y_n)\}$, where $y_i=1$ for $i={n^-+1},\ldots,n$. When we want to emphasize that an input belongs to $D^{-}$ we denote that by $\mathbf{x}^{-}_i$, for $i=1,\ldots,n^-$, and similarly if $(\mathbf{x}_{i},y_{i})\in D^{+}$ we may denote the input by $\mathbf{x}^+_{i}$, for $i=n^-+1,n^-+2,\ldots, n$. We follow the convention that names one dataset, $D^-$, negative and the other, $D^+$, positive. However, we label the data points $0$ for the former, and $1$ for the latter which facilitates further probabilistic formulations. 

Next, define a similarity function. Following~\citep{gilboa2006empirical}, parametrize the similarity function with a vector $\mathbf w\in\mathbb{R}^{+p}$. The role of $\mathbf w$ is to define a weighted distance between $\mathbf x=[x_1,\ldots,x_p]^T$ and $\mathbf x'=[x'_1,\ldots,x'_p]^T\in\mathbb{R}^p$, specifically, 
\begin{eqnarray}
d_{\mathbf w}= \sqrt{\sum_{j=1}^p w_jd_j^2},\label{eq:distance}
\end{eqnarray}
where $\mathbf w=[w_1,\ldots,w_p]^T$, and $d_j = x_j-x'_j$, for $j=1,\ldots,p$. Then define the similarity function
\begin{eqnarray}
S_{\mathbf w}= \exp{\left\{-d_\mathbf{w}\right\}},\label{eq:similarity}
\end{eqnarray}
where $d_\mathbf{w}$ is defined in~\eqref{eq:distance}. In fact, $S_{\mathbf w}$ assigns a higher similarity measure to the points that have a smaller weighted distance with each other and a lower similarity measure to the points that are not close to each other based on the metric $d_{\mathbf w}$. Intuitively, a large value for a component $w_j$ implies that the corresponding input dimension $x_j$ contributes more to the value of the similarity function. Specifically, if $w_j>w_k$, a unit of increase in the direction of the $j$th component (i.e. changing $d_j$ to $d_j+1$) will reduce $S_{\mathbf{w}}$ more, compared to that for the $k$th component, assuming $d_j=d_k$. While other similarity functions can also be employed in this framework, formulas~\eqref{eq:distance} and~\eqref{eq:similarity} are equivalent to a set of consistency conditions on the response $y$ (see~\citep{billot2008axiomatization} for details of this axiomatization).

Next, write the weighted average of the data points based on the similarity function $S_{\mathbf w}$,
\begin{eqnarray}
z_i=\frac{\sum_{\ell\neq i} S_{\mathbf w}(\mathbf x_i,\mathbf x_{\ell})y_{i}}{\sum_{\ell\neq i} S_{\mathbf w}(\mathbf x_i,\mathbf x_{\ell})},\label{eq:zt}
\end{eqnarray}
which is always between $0$ and $1$. A more general form which allows for more complicated relationship between $z_i$ and $P(y_i=1|\mathbf x_i,D\backslash (\mathbf x_i,y_i))$ can also be used, such as any cumulative distribution function (CDF) whose support is the set $[0,1]$ to relate the probability of $y_i=1$ to $z_i$, i.e.,
\begin{eqnarray}
P(y_i=1|\mathbf x_i,D\backslash (\mathbf x_i,y_i))=F(z_i),\label{eq:F}
\end{eqnarray}
where $F(z_i)$ denotes the value of the CDF, $F$, evaluated at $z_i$. Note that since $F$ is non-decreasing, a higher value for $z_i$ shows a higher probability for $\mathbf x_i$ belonging to the positive class. To find optimal values for $\mathbf w$, maximize the log-likelihood function
\begin{eqnarray}
l=\sum_{i=1}^n\left\{Y_i\ln(F(z_i))+(1-Y_i)\ln(1-F(z_i))\right\}.\label{eq:log_lik_unc}
\end{eqnarray}
Recall that if the majority of data points in $D$ belong to the negative class, i.e., if $n^+\ll n^-$, classification algorithms generally label many of the test points belonging to the positive class incorrectly, i.e. negative~\citep{He:09}. Note, too, that the optimized similarity function will be biased towards labeling most test points as negative, even though they may belong to the positive class, if the dataset is overwhelmed by one class. For some data structures, the poor performance of classification techniques can be attributed to insufficient information as a result of too few data points in $D^{+}$. Note that ``creating'' extra synthetic data points using the current dataset may improve algorithmic performance, but doing so may introduce bias. The next sub-section explains how absent data generation may provide an acceptable balance between the expected classification error and the bias. 
\subsection{Absent data generation}\label{sub_sec_absent}

As mentioned in Section~\ref{sec_intro}, for many imbalanced classifications, the crucial property of an algorithm is its ability to correctly identify the test points belonging to the minority class. Therefore, balancing the dataset by generating the synthetic data points belonging to the minority  class enhances the algorithm’s detection power~\citep{Chawla:02}.  Most synthetic data generation methods have two  independent algorithms: one that creates synthetic data, and one that performs classification on the new dataset consisting of the actual and synthetic data points.  If the synthetic data generation mechanism is embedded within  the classification algorithm, the new data points are generated such that  the performance of the classifier improves  compared to having a data generation algorithm independent from the classification algorithm~\citep{pourhabib2015absent}. Hence, the idea of absent data comes into play, i.e., the data points that, if they existed, would help the classification algorithm better identify the test points belonging to the minority class. Absent data can be considered  as a special case of synthetic data whose properties can be used to improve an algorithm’s detection power  without needing to generate the synthetic data.

Let $\mathbf x^a_t\in\mathcal{X}^+$, for $t=1,\ldots,T$ denote absent data points, where $\mathcal{X}^+\subset\mathcal{X}$ represents the input domain for minority inputs and use the points to construct constraints that mitigate the low detection power problem. Since the absent data points compensate for a lack of sufficient number of minority points, they need to belong to the minority domain $\mathcal{X}^+$. To ensure each $\mathbf x^a_t$ belongs to $\mathcal{X}^+$, restrict  the absent data points to be ``close'' to the existing minority points in $D^+$. To define closeness, employ the similarity function to make sure the absent points are similar, determined by function $S_{\mathbf w}(\cdot,\cdot)$, to the existing minority points, i.e.,   
\begin{eqnarray}
\sum_{t=1}^T\sum_{i=n^{-}+1}^{n} S_{\mathbf w}(\mathbf x^{+}_i,\mathbf x^a_{t})\ge\Delta, \label{eq:const_min}
\end{eqnarray}
for some $\Delta>0$. Constraint~\eqref{eq:const_min} states that the overall similarity of all of the absent data points  to the existing minority data points should exceed some threshold. 

Absent data being similar to the existing minority data, however, does not guarantee their usefulness. In other words, the synthetic data are useful as long as they are close to the boundary~\citep{Han:05}and the absent data must not be far away from the existing majority points, specifically,
\begin{eqnarray}
\sum_{t=1}^T\sum_{i=1}^{n^{-}} S_{\mathbf w}(\mathbf x^{-}_i,\mathbf x^a_{t})\ge\delta, \label{eq:const_maj}
\end{eqnarray}
for some $\delta>0$. Constraint~\eqref{eq:const_maj} may appear counter-intuitive, but recalling the role of absent data, which is to facilitate the correct boundary identification, leads to the realization that the data points residing far from the boundary between the two classes will not be informative.  In fact, constraints~\eqref{eq:const_min} and~\eqref{eq:const_maj} together enforce that the absent data points fall in a region separating the two classes. It is preferable to use constraints~\eqref{eq:const_min} and~\eqref{eq:const_maj} to address the overall similarity between all absent data points and existing majority/minority data points rather than enforcing similarity between all individual points, because the latter approach makes likelihood maximization very challenging due to the resulting large number of constraints.

Therefore, maximize the log-likelihood~\eqref{eq:log_lik_unc} subject to constraints~\eqref{eq:const_min} and~\eqref{eq:const_maj},
\setlength{\arraycolsep}{0.0em}
\begin{eqnarray}
\max l=\notag\\
&&\sum_{i=1}^n\left\{Y_i\ln(F(z_i))+(1-Y_i)\ln(1-F(z_i))\right\},\label{eq:log_lik}\\
\text{s.t.}\notag\\
&&\sum_{t=1}^T\sum_{i=n^{-}+1}^{n} S_{\mathbf w}(\mathbf x^{+}_i,\mathbf x^a_{t})\ge\Delta,\notag\\
&&\sum_{t=1}^T\sum_{i=1}^{n^{-}} S_{\mathbf w}(\mathbf x^{-}_i,\mathbf x^a_{t})\ge\delta,\label{eq:log_lik_con}
\end{eqnarray}
\setlength{\arraycolsep}{5pt}
for given $\delta>0$, and $\Delta>0$, where the decision variables are $\mathbf w$, and $\mathbf x^a_{t}$, $t=1,\ldots,T$. To solve optimization problem~\eqref{eq:log_lik_con}, write the Lagrangian of the problem as,   
\begin{eqnarray}
\max g(\mathbf w,\mathbf X^a)=&&l+\lambda_1(\sum_{t=1}^T\sum_{i=n_{-}+1}^{n} S_{\mathbf w}(\mathbf x^{+}_i,\mathbf x^a_{t})-\Delta)\notag\\
&&+\lambda_2(\sum_{t=1}^T\sum_{i=1}^{n_{-}} S_{\mathbf w}(\mathbf x^{-}_i,\mathbf x^a_{t})-\delta),\label{eq:Lag}
\end{eqnarray}
where $\lambda_1\ge 0$ and $\lambda_2\ge 0$ are the Lagrangian coefficients, and $\mathbf X^a$ is an $p\times T$ matrix whose $t^{\text{th}}$ column is $\mathbf x^a_t$. $l$ was defined in~\eqref{eq:log_lik_unc}. 

It is possible to interpret optimization problem~\eqref{eq:Lag} as a penalized log-likelihood maximization, specifically, by the weights  $\mathbf w$ that  maximize the likelihood and penalizing any violation of the constraints related to the absent data. Assume the Lagrangian coefficients are given and find the stationary points of the objective function $g(\mathbf w,\mathbf X^a)$ in~\eqref{eq:Lag},
\begin{eqnarray}
\frac{\partial g}{\partial w_j}=&&\sum_{i=1}^n\frac{(Y_i-F(z_i))f_i}{F(z_i)(1-F(z_i))}\frac{\partial}{\partial w_j}z_i\notag\\
&&+\lambda_1\sum_{t=1}^T\sum_{i=n^{-}+1}^{n} \frac{\partial}{\partial w_j}S_{\mathbf w}(\mathbf x^{+}_i,\mathbf x^a_{t})\notag\\
&&+\lambda_2\sum_{t=1}^T\sum_{i=1}^{n^{-}} \frac{\partial}{\partial w_j}S_{\mathbf w}(\mathbf x^{-}_i,\mathbf x^a_{t}),\label{eq:parder_w}
\end{eqnarray}
for $j=1,\ldots,m$, where $f_i$ is the probability distribution function of $F(z_i)$. Note that 
 \begin{eqnarray}
\frac{\partial }{\partial w_j}z_i=\frac{A\sum_{\ell\neq i}\dot{S}_{\mathbf w,j}(\mathbf x_i,\mathbf x_{\ell})y_i-AY\sum_{\ell\neq i}\dot{S}_{\mathbf w,j}(\mathbf x_i,\mathbf x_{\ell})}{A^2}
\end{eqnarray} 
where 
 \begin{eqnarray}
&& A=\sum_{\ell\neq i}S_{\mathbf w}(\mathbf x_i,\mathbf x_{\ell}),\notag\\
&& AY=\sum_{\ell\neq i}S_{\mathbf w}(\mathbf x_i,\mathbf x_{\ell})y_i,\notag\\
\dot{S}_{\mathbf w,j}(\mathbf x_i,\mathbf x_{\ell})&&=\frac{\partial}{\partial w_j}S_{\mathbf w}(\mathbf x_i,\mathbf x_{\ell})\notag\\
&&=-\frac{S_{\mathbf w}(\mathbf x_i,\mathbf x_{\ell})(x_{ij}-x_{\ell j})^2}{2d_\mathbf{w}(\mathbf x_i,\mathbf x_\ell)}.
\end{eqnarray} 

The partial derivatives of $g$ with respect to the absent data points are
\begin{eqnarray}
\frac{\partial g}{\partial x^a_{tj}}=&&\lambda_1\sum_{i=n^{-}+1}^{n} \frac{S_{\mathbf w}(\mathbf x^{+}_i,\mathbf x^a_{t})\times w_j\left(x_{ij}-x^a_{tj}\right)}{d(\mathbf x^{+}_i,\mathbf x^a_{t})}\notag\\
&&+\lambda_2\sum_{i=1}^{n^{-}} \frac{S_{\mathbf w}(\mathbf x^{-}_i,\mathbf x^a_{t})\times w_j\left(x_{ij}-x^a_{tj}\right)}{d(\mathbf x^{-}_i,\mathbf x^a_{t})},\label{eq:parder_abs}
\end{eqnarray}
for $j=1,\ldots,p$, and $t=1,\ldots,T$, where $\mathbf x^a_t=[x^a_{t1},\ldots,x^a_{tp}]^T$, and $\mathbf x_i=[x_{i1},\ldots,x_{ip}]^T$.

Solve the total of $(T+1)p$ equations
\begin{eqnarray}
&&\frac{\partial g}{\partial x^a_{tj}}=0,\text{ for } j=1\ldots,p, t=1,\ldots,T\label{eq:fonc_x}\\
&&\frac{\partial g}{\partial w_j}=0,\text{ for } j=1\ldots,p,\label{eq:fonc_w} 
\end{eqnarray}
using iterative numerical techniques, such as a trust region algorithm~\citep{byrd2000trust,conn2000trust} to minimize the sum of squares of $\frac{\partial g}{\partial x^a_{tj}}$ and $\frac{\partial g}{\partial w_j}$, which can be conducted in polynomial time. Since the solution to equations~\eqref{eq:fonc_x}-\eqref{eq:fonc_w} are the points that satisfy the first-order necessary conditions, which due to the non-convexity of~\eqref{eq:Lag} are not necessarily the global optimal points of optimization problem~\eqref{eq:Lag}, note that the proposed algorithm may become trapped in local optima for some datasets.  

Last, we need to determine the values of the Lagrangian coefficients $\boldsymbol{\lambda}=[\lambda_1,\lambda_2]^T$. The Lagrangian relaxation provides an upper bound for the original problem. To obtain the solution of~\eqref{eq:log_lik}~\eqref{eq:log_lik_con}, minimize the maximum value of the Lagrangian relaxation. Specifically,
\begin{eqnarray}
&&\min R(\boldsymbol{\lambda})\text{ s.t. } \boldsymbol{\lambda}\geq \boldsymbol{0},\label{eq:minR}
\end{eqnarray}
where $R(\boldsymbol{\lambda})$ is the value of the objective function in~\eqref{eq:Lag}, i.e., for a given $\boldsymbol{\lambda}$, if $(\tilde{\mathbf w},\tilde{\mathbf X}^a)$ is a solution to ~\eqref{eq:fonc_x} and~\eqref{eq:fonc_w}, $R(\boldsymbol{\lambda})= g(\tilde{\mathbf w},\tilde{\mathbf X}^a)$. Section~\ref{sub_sec_param} provides a discretization scheme to approximate $R(\boldsymbol{\lambda})$, because solving~\eqref{eq:minR} to optimality  is challenging.

\subsection{Cluster-based undersampling}

Combining undersampling of the majority data points with oversampling (synthetic or actual) of the minority data points~\citep{Chawla:02} helps to identify the correct boundary in imbalanced data structures.  Efficiency is another reason to conduct undersampling, since a large number of majority points slows the iterative procedure for solving equations~\eqref{eq:fonc_x} and~\eqref{eq:fonc_w},

Let $D_{\mathbf x}^-$ denote the inputs in the training dataset containing the majority points. That is, $D_{\mathbf x}^-=\{\mathbf x_1,\mathbf x_2,\ldots,\mathbf x_n^-\}$ such that $(\mathbf x_i,y_i)\in D^-$, for $i=1,\ldots,n^-$. Cluster $D_{\mathbf x}^-$ into $K\in\mathbb{N}$ clusters, $\{C_1,\ldots,C_K\}$, where $C_{\ell}\cap C_k=\phi$, for $\ell\neq k$, and $\bigcup C_{\ell}=D_{\mathbf x}^-$. Then, for every $\mathbf x_i\in D_{\mathbf x}^-$, there exists one (and only one) $C_{\ell}$ such that $\mathbf x_i\in C_{\ell}$. Create $U\in\mathbb{N}$ undersampled majority training datasets $\{D^-_1,\ldots,D^-_U\}$ such that every $D^-_{\ell}$, $\ell=1,\ldots,U$ contains $K$ majority training data points. Specifically, 
\begin{eqnarray}
D^-_{\ell}=\{(\mathbf x_{i_1},y_{i_1}),(\mathbf x_{i_2},y_{i_2}),\ldots,(\mathbf x_{i_K},y_{i_K})\},\label{eq:data_clus}
\end{eqnarray}
where $\mathbf x_{i_j}\in C_{j}$, for $j=1,\ldots,K$. In other words, each $D^-_{\ell}$ contains $K$ data points, where the input for each data point comes from one of the $K$ clusters $\{C_1,\ldots,C_K\}$. To create each $D_{\ell}$ perform random undersampling. 

Then train the model $U$ times based on the undersampled dataset $D_\ell:=(D^-_\ell,D^+)$, for $\ell=1,\ldots,U$, specifically use $D_\ell$ to solve~\eqref{eq:fonc_x} and~\eqref{eq:fonc_w}. Each of these trainings provides an estimate for the probability of the training points being one, i.e. $\widehat{p}_\ell(\mathbf x_*)=P(y_*=1|D_{\ell},\mathbf x_*)$, where $\mathbf x_*\in \mathcal{X}$ is a test point. The sample average of all estimates serves as the prediction of the probability $P(y_*=1|\mathbf x_*)$. Such ensemble learning~\citep{hastie:09} based on undersampled majority data points has proven powerful in handling imbalanced data structures~\citep{liu2009exploratory}. We use a $k$-means algorithm to cluster the majority inputs, which can be implemented in a close to linear time complexity~\citep{kanungo2002efficient}. Section~\ref{sub_sec_param} presents guidelines for selecting the number of clusters for each dataset. Based on the proposed framework, the algorithmic steps are as follows (Table~\eqref{SBIC_alg} lists the steps). 

Let $(\mathbf w_\ell,\mathbf X^a_\ell)$ denote the stationary points for $g(\mathbf w,\mathbf X^a)$ based on the training data in $D_\ell$, i.e., instead of using all the points in $D$, use the smaller set $D_{\ell}$ to solve~\eqref{eq:fonc_x} and~\eqref{eq:fonc_w}. 
Use $\mathbf w_\ell$ to calculate $z_i$ in~\eqref{eq:zt}, and also use it to calculate $P(y_*=1|\mathbf x_*,D_\ell)$, i.e., the probability of a test point $\mathbf x_*$ belonging to the minority class based on the dataset $D_\ell$. Find the probability $P(y_*=1|\mathbf x_*)$ by averaging over all the predicted probabilities, specifically,
\begin{eqnarray}
P(y_*=1|\mathbf x_*) = \sum_{\ell=1}^U{\pi(D_\ell)P(y_*=1|\mathbf x_*,D_\ell)},
\end{eqnarray}
where $\pi(D_\ell)$ is the prior probability associated with the dataset $D_{\ell}$. Assigning equal prior probability to each dataset yields
\begin{eqnarray}
P(y_*=1|\mathbf x_*) = \frac{1}{U}\sum_{\ell=1}^U{P(y_*=1|\mathbf x_*,D_\ell)}.\label{eq:pred_prob}
\end{eqnarray}

We call the proposed approach for estimating $P(y_*=1|\mathbf x_*)$ Similarity-based Imbalanced Classification (SBIC). In SBIC, although the values of absent data points that solve equations~\eqref{eq:fonc_x} do not appear in~\eqref{eq:pred_prob}, they impact the optimal values of $\mathbf w$ which determines $z_*$ in equation~\eqref{eq:zt}. In other words, incorporating absent data  into the formulation guides the similarity weights, $\mathbf w$, to self-adjust themselves, as though the absent data actually exist. Notably, SBIC simultaneously both absent data points and similarity weights simultaneously, and then uses the latter for prediction. 

\begin{algorithm}
\caption{Similarity-based Imbalanced Classification}
\label{SBIC_alg}
\begin{algorithmic}
\STATE 1. {Given $D^-$, $D^+$, $K$, $U$, $T$, $F(\cdot)$, $\boldsymbol{\lambda}^d$, $\delta$, $\Delta$, $\mathbf x_*\in\mathcal{X}$.}

\STATE 2. {Cluster $D^-$ into $K$ sets.  Let $\ell=1$, $P=0$.}
\REPEAT{

	\STATE 3. {Create $D^{-}_\ell$ according to~\eqref{eq:data_clus}. Let $D_\ell=D^{-}_\ell\cup D^+$.} 
	
	\STATE 4. {Let $\boldsymbol{\lambda}=\boldsymbol{\lambda}^d(\ell).$}
					
			\STATE 5. {Let $(\mathbf w_\ell,\mathbf X_\ell^a)$ be a solution to ~\eqref{eq:fonc_x} and~\eqref{eq:fonc_w} based on the data points in $D_\ell$.}

			\STATE 6. {Calculate $S_{\mathbf w_\ell}$ according to~\eqref{eq:similarity}.}

			\STATE 7. {$z_*=\frac{\sum_{\ell\neq i} S_{\mathbf w_\ell}(\mathbf x_i,\mathbf x_{\ell})y_{t}}{\sum_{\ell\neq i} S_{\mathbf w_\ell}(\mathbf x_i,\mathbf x_{\ell})}$}

			\STATE 8. {$P=P(y_*=1|\mathbf x_*,D_\ell) = F(z_*)$}

	\STATE 9. {$\ell=\ell+1$.}
}
\UNTIL{$\ell>U$}.

\STATE 10. $P(y_*=1|\mathbf x_*)=\frac{1}{U}P$

\end{algorithmic}
\end{algorithm}

Algorithm~\ref{SBIC_alg} assumes the values of the Lagrangian coefficients $\boldsymbol{\lambda}=[\lambda_1,\lambda_2]^T$ are given. Section~\ref{sub_sec_param} discusses how we obtain the Lagrangian coefficients. In the algorithm, $\boldsymbol{\lambda}^d$ stores all the values of $\boldsymbol{\lambda}$ for any $D_\ell$, and the algorithm picks the associated value by assigning $\boldsymbol{\lambda}^d(\ell)$ to $\boldsymbol{\lambda}$.

\section{Numerical studies}\label{sec_numerical}

Comparing the performance of different algorithms on imbalanced datasets requires care. Section~\ref{sub_sec_eval_crit} gives the details, SectionSection~\ref{sub_sec_param} discusses the selection of parameters assumed given for Algorithm~\eqref{SBIC_alg}, and Section~\ref{sub_sec_results} compares SBIC with competing algorithms.

\subsection{Evaluation Criteria}\label{sub_sec_eval_crit}

In general, for a two-class classification, an algorithm is deemed effective if it can  correctly label test points as positive or negative. If $D_*=\{(\mathbf{x}_{*1},y_{*1}),(\mathbf{x}_{*2},y_{*2}),\dots,(\mathbf{x}_{*n_{te}},y_{*n_{te}})\}$ denotes the test dataset, and $\widehat{y}_i$ denotes the predicted label for the test input $i=1,\ldots,n_*$, then measure
\begin{eqnarray}
E=\frac{1}{n_{te}}\sum_{i=1}^{n_{te}}I(y_{*i}=\widehat{y}_i),\label{eq:error}
\end{eqnarray}
where $I(\cdot)$ is an indicator function which returns $1$ if its argument is true. However, in most classification applications, particularly for imbalanced classifications, the cost associated with incorrectly labeling the positive points as negative is much higher than the opposite. Therefore, it is important to distinguish between the two types of error, false alarm and mis-detection. Specifically, let $D^+_*\subset D_*$ and $D^-_*\subset D_*$ denote the subset of the test dataset that contain the positive and negative labels, respectively. Define false alarm
\begin{eqnarray}
FA=\frac{1}{|D^-_*|}\sum_{i}I(y_{*i}=\widehat{y}_i),\text{ for }(\mathbf{x}_{*i},y_{*i})\in D^-_*,\label{eq:FA}
\end{eqnarray}
where $|D^-_*|$ denotes the number of negative points in the test dataset. Now define mis-detection
\begin{eqnarray}
MD=\frac{1}{|D^+_*|}\sum_{i}I(y_{*i}=\widehat{y}_i),\text{ for }(\mathbf{x}_{*i},y_{*i})\in D^+_*.\label{eq:MD}
\end{eqnarray}
Ideally, $FA=MD=0$, but it does not happen except for trivial cases. Also note that SBIC is a probabilistic classifier. That is, SBIC does not directly predict positive or negative labels for a given test point $\mathbf x_*$, but it does provide a probability $P(y_*=1|\mathbf x_*)$, also called a score, as noted in equation~\eqref{eq:pred_prob}. Therefore, assign a label to $\mathbf x_*$ by defining a decision threshold between $0$ and $1$, where a test point with $P(y_*=1|\mathbf x_*)$ less than or equal to the threshold is labeled negative. Changing the decision threshold can give different values for $FA$ and $MD$. A trade-off between false alarm and mis-detection implies that reducing the threshold increases false alarms and decreases mis-detections. 

The receiver operating characteristic curve, (ROC curve)  formalizes the idea of evaluating a probabilistic classifier by changing the decision threshold~\citep{bradley1997use}.  The details are as follows. In an ROC curve, the $x$-axis denotes the false alarm and the $y$-axis denotes 1−mis-detection,  also called the detection power (DP). Setting the decision threshold to $1$ corresponds to the point (0, 0), i.e., a classifier with no FA and no DP. Gradually reducing the threshold with steps smaller than the minimum value of the differences between scores yields a point with either a higher FA or a higher DP, and continuing to do so yields points on the ROC space, with each representing a (FA,DP)  combination. Connecting all points yields the ROC curve.

\begin{figure}[h]
\centering
		\includegraphics[width=0.5\textwidth]{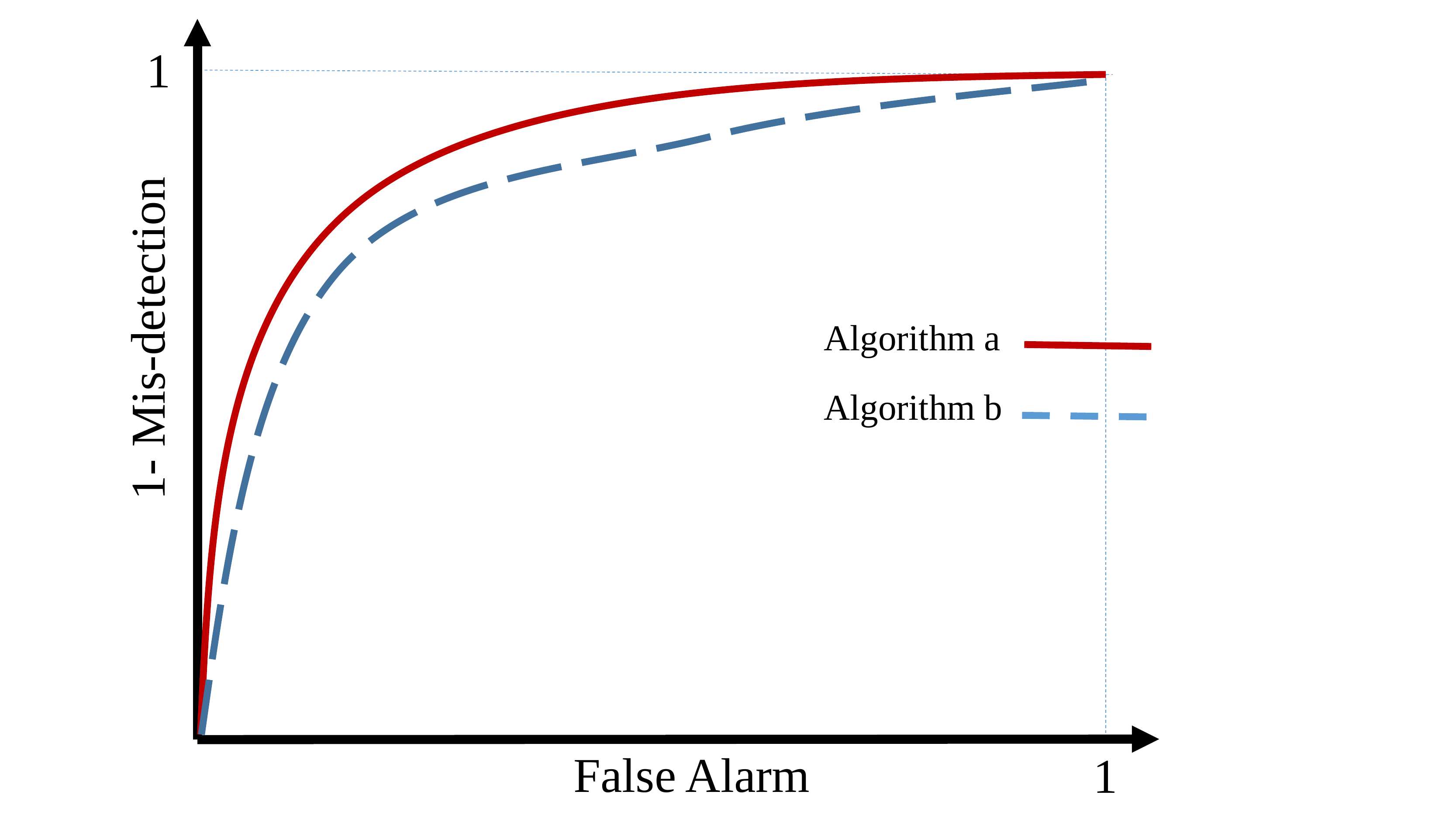}
		\caption {The ROC curve for two algorithms. Algorithm a is superior since its associated area under curve (AUC) is larger than that of Algorithm b.}
\label{fig_roc_illus}
\end{figure}

An algorithm is deemed relatively superior when its ROC curve is close to the two-segment line from $[0, 0]$ to $[0, 1]$ and from $[0, 1]$ to $[1, 1]$ on the FA-DP axes, i.e., the curve is closer to the top-left region of the plot (see Figure~\ref{fig_roc_illus}). To evaluate performance, simply measure the area under the ROC curve, or AUC. An algorithm with a larger AUC, i.e., closer to $1$, is deemed superior for a given dataset. Once an ROC curve is generated, use numerical integration to calculate the corresponding AUC.

\subsection{Parameters of SBIC}\label{sub_sec_param}

Algorithm~\ref{SBIC_alg} has a set of user-defined parameters. This section gives some guidelines for their selection. 

The number of absent data points, $T$ , impacts the optimal values of the weights $\mathbf w$ as well as the efficiency of the model, since the number of equations to obtain the stationary points is $(T + 1)p$. Generating synthetic data balances the dataset, and there is no need to generate absent data. In fact the role of absent data is to guide the weights in order to account for the dataset's imbalanced structure. Thus, a large number of absent data points does not necessarily improve the algorithm’s prediction capability. The following implementation uses $T = p$, i.e., the dimension of the input space $X$ . Based on the experimental results, this setting provides a good balance between prediction accuracy and efficiency.

Lagrangian coefficients $\boldsymbol{\lambda}=[\lambda_1,\lambda_2]^T$ in~\eqref{eq:Lag} determine how much to penalize violations of constraints~\eqref{eq:const_min} and~\eqref{eq:const_maj}. Note that in Algorithm~\ref{SBIC_alg} each undersampled data $D_\ell$ needs a value for $\boldsymbol{\lambda}$. Perform an exhaustive search to  obtain the optimal value for $\boldsymbol{\lambda}$. Specifically, let $\Lambda_1=\{\lambda_{11}^c,\lambda_{12}^c,\ldots,\lambda_{1n_c}^c\}$ denote the set of candidate values for $\lambda_1$, and $\Lambda_2=\{\lambda_{21}^c,\lambda_{22}^c,\ldots,\lambda_{2n_c}^c\}$ denote the set of candidate values for $\lambda_2$. Then, for a given $D_\ell$ solve equation~\eqref{eq:fonc_x} and~\eqref{eq:fonc_w} and evaluate 
\begin{eqnarray}
R (\lambda_{1p}^c,\lambda_{2q}^c)=&&l+\lambda_{1p}^c(\sum_{t=1}^T\sum_{i=n_{-}+1}^{n} S_{\mathbf w}(\mathbf x^{+}_i,\mathbf x^a_{t})-\Delta)\notag\\
&&+\lambda_{2q}^c(\sum_{t=1}^T\sum_{i=1}^{n_{-}} S_{\mathbf w}(\mathbf x^{+}_i,\mathbf x^a_{t})-\delta),\label{eq:R} 
\end{eqnarray}
for $p,q\in\{1,\ldots,n_c\}$. Store the optimal values based on this approach in the array $\boldsymbol{\lambda}^d$, which is used in Algorithm~\ref{SBIC_alg}. Note that such a discretization approach does not provide the optimal solution to~\eqref{eq:minR}; however, for any values of $(\lambda_1,\lambda_2)$, a solution to optimization problem ~\eqref{eq:minR} provides an upper bound for the optimization problem~\eqref{eq:log_lik}-\eqref{eq:log_lik_con}. Now utilize $\mathbf w_*$ that results in an upper bound for~\eqref{eq:log_lik}-\eqref{eq:log_lik_con} to make a prediction at $x_*\in\mathcal{X}$.

Parameters $\Delta$ and $\delta$ in~\eqref{eq:const_min} and~\eqref{eq:const_maj} determine a threshold for the similarity between the absent points and the minority or  majority points, respectively. These parameters only appear in finding the optimal values for $\boldsymbol{\lambda}$ in~\eqref{eq:R}. We suggest 
\begin{eqnarray}
&&\Delta = \frac{T}{n^+}\sum_{i<j}^{n} S_{\mathbf w}(\mathbf x^{+}_i,\mathbf x^{+}_j), \text{ for }\mathbf x^{+}_i,\mathbf x^{+}_j\in D^+,\label{eq:BDelta}\\
&&\delta = \frac{T}{4|D^-_\ell|}\sum_{i<j}^{n} S_{\mathbf w}(\mathbf x^{-}_i,\mathbf x^{-}_j), \text{ for } \mathbf x^{-}_i,\mathbf x^{-}_j\in D^-_\ell\label{eq:Sdelta}.
\end{eqnarray}
Equation~\eqref{eq:BDelta} implies that the average similarity between an absent data point and the minority data point should be greater than that between the minority points themselves. Equation~\eqref{eq:Sdelta} implies that the average similarity between an absent data point and the majority point should be greater than $25\%$ of the similarity between the majority points. Both equations use a $\mathbf w$ that is a local optimum of the likelihood function in~\eqref{eq:log_lik_unc}. 

Recall that clustering dataset $D$ into $K$ clusters  in order to build an ensemble learner and to improve the efficiency of solving equations~\eqref{eq:fonc_x} and~\eqref{eq:fonc_w}. In other words, $K$ needs to be small enough to have a sufficient number of data points in each $D_\ell$ and large enough to efficiently solve~\eqref{eq:fonc_x} and~\eqref{eq:fonc_w}. If some data points are densely aggregated in one region, consider it as one cluster which in turn reduces $K$. Therefore, the selection of $K$ depends on the specific dataset. In this implementation  $K$ is selected to balance the relative size of the majority to minority points, and $K$ is always greater  than $50$ to maintain the effectiveness of each $D_\ell$. $K$ relates to $U$, i.e. the number of undersampled datasets. If the number of data points in each cluster is small, a small value for $U$  is sufficient, whereas if the number of data points is large, a larger $U$ is needed. In this implementation, depending on the dataset, $K$ ranges between $1$, i.e., only one undersampled dataset for model training, and $10$.

Finally, we need to determine the CDF, $F(\cdot)$, in~\eqref{eq:F}. Note that $z_i$ in~\eqref{eq:zt} is always between $0$ and $1$, i.e., the support for the CDF should be between $0$ and $1$. Therfore, use distributions such as beta or uniform. In this implementation the uniform distribution for $F$, specifically, $F(z_i)=z_iI(0\le z_i\le 1)$ is selected.

\subsection{Toy examples}

\begin{figure*}[h]
\centering
		\includegraphics[width=1\textwidth]{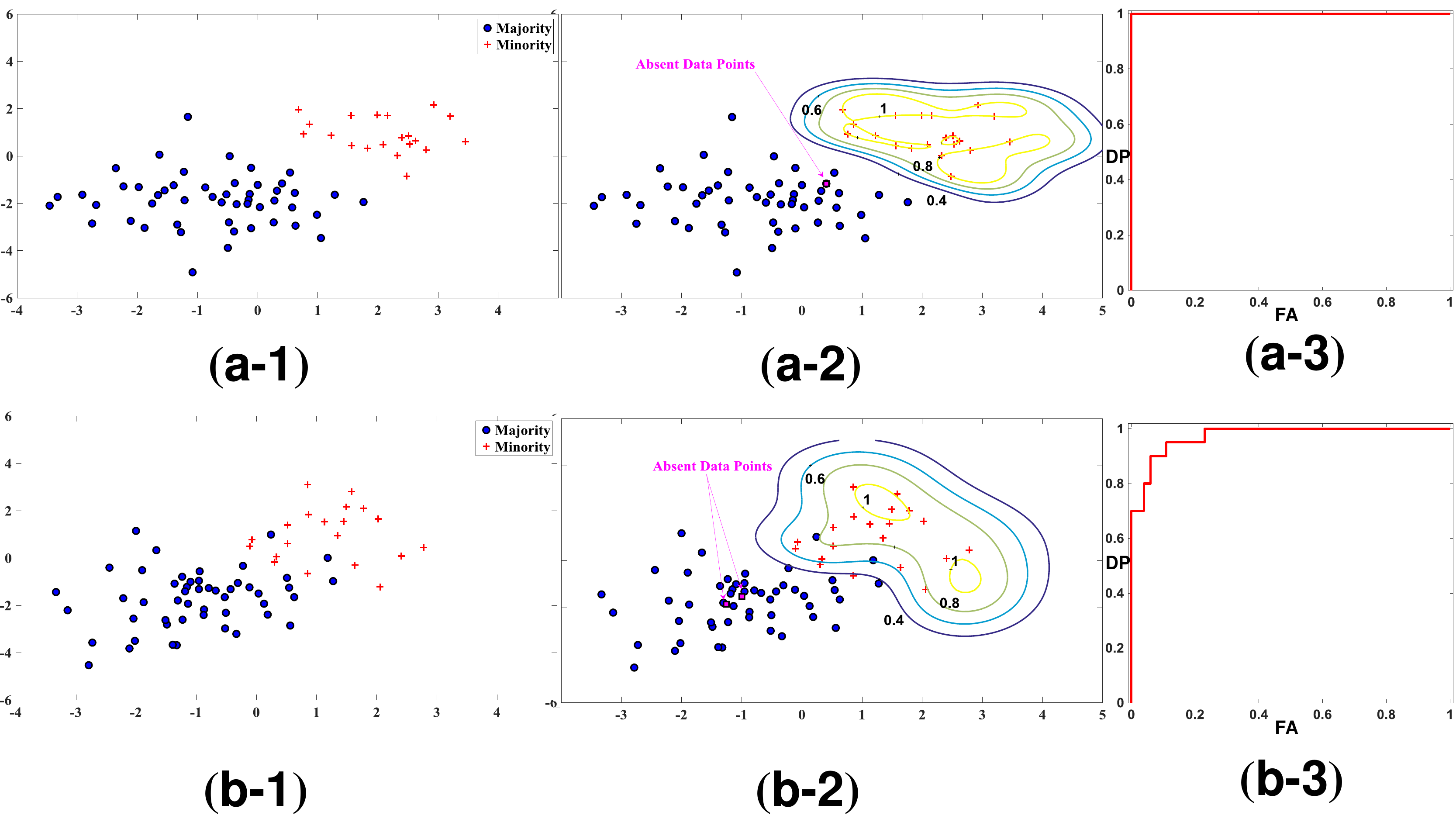}
		\caption {Top: (a-1) Dataset \texttt{Toy1} which has two separate classes. (a-2) The locations of absent data points (two points very close to each other) and the contour plots for the predicted probabilities:  the numbers on the contour curves denote the probability of belonging to the minority class. (a-3) The ROC curve for  \texttt{Toy1}. Bottom: (b-1)  Dataset \texttt{Toy2} which has two overlapping classes. (b-2) The locations of absent data points and the contour plots for the predicted probabilities: the numbers on the contour curves denote the probability of belonging to the minority class. Compared with plots (a), the absent data points are distinct and deeper into the majority region. (b-3) The ROC curve for  \texttt{Toy2}.}
\label{fig_toy_rev}
\end{figure*}

Before reporting the results on real datasets,  we present the performance of SBIC on the following three simulated datasets. We generate $n^-=100$ data points from a normal distribution  with mean  $[-1, -2]^T$  and variance-covariance matrix   $[1.1, 0.1; 0.1, 1.2]$, which constitute the majority  data points.  For the minority data points, we create $n^+=20$  samples from a normal distribution with mean  $[2, 1]^T$ and variance-covariance matrix $[0.6, -0.1; -0.1, 1.7]$. \texttt{Toy1}, therefore, is the dataset with well-separated minority and majority samples (see plot (a-1) in Figure~\ref{fig_toy_rev}).  We create another set of samples from a normal distribution with mean $[1, 1]^T$ and variance-covariance matrix $[0.6, -0.1; -0.1, 1.7]$. \texttt{Toy2}, therefore, is the dataset with aggregated minority  and majority samples (see plot (b-1) in Figure~\ref{fig_toy_rev}). Note that for both \texttt{Toy1} and \texttt{Toy2}, we undersample the majority datasets  so that we have  $n^-=50$ remaining majority points. The plots in ~\ref{fig_toy_rev} only depict the $50$ majority points along with the original $n^+=20$ minority  points. Both toys use $T=2$ absent data points in the implementation of SBIC.

Plot (a-2) in Figure~\ref{fig_toy_rev} shows the locations of the absent data points found by SBIC algorithm~\ref{SBIC_alg}, and the contour plots of the probabilities of belonging to the minority class. We obtain the contour plots by creating test points close to the minority training samples and then fit a continuous surface to the estimated probabilities obtained through SBIC algorithm~\ref{SBIC_alg}. The numbers on each contour curves denotes the probability of belonging to the minority class. Observing that the absent data points are both at the same location suggests that when the samples from two classes are well separated and we have a relatively sufficient number of training samples, the absent data points do not play an important role. Plot (a-3) in Figure~\ref{fig_toy_rev} shows the ROC curve for this example, which has a corresponding AUC=$99.99\%$.

When the minority and majority regions have more overlaps, the absent data points significantly impact SBIC algorithm~\ref{SBIC_alg}. Plot (b-2) in Figure~\ref{fig_toy_rev} shows that the locations of the absent data  points are close to the boundary of the two classes, but compared to \texttt{Toy1}, they are further inside the majority region. Loosely speaking, the absent data points try to explore the majority region so that they are positioned in an area that helps the algorithm to better identify the boundary. Again, we note an important difference between synthetic data points in general and absent data points: the former represents the data points from the minority class, whereas the latter helps the algorithm to identify the minority region. As such, the locations of the absent data points would not necessarily be the same as the locations of extra samples possibly obtained from the minority class by linear interpolation~\citep{Chawla:02}, but they are parameters in optimization problem~\eqref{eq:log_lik_con}. We adjust these parameters to optimize the algorithm's overall detection power. As mentioned  we use the values of weights $\mathbf w$ rather than utilizing the actual values of the optimal absent data points in prediction. While the value of the absent  data points affects optimal $\mathbf w$, i.e.  a solution to optimization problem (11), the deep intrusion of absent data points into the majority  region for \texttt{Toy2} violates the idea of having $\mathbf x^a_t\in\mathcal{X}^+$ as discussed  in Section~\ref{sub_sec_absent}. This can be a result of solving the relaxation
of optimization problem~\eqref{eq:log_lik_con}. Our discretization approach to find $\lambda_1$ and $\lambda_2$ for optimization problem~\eqref{eq:minR}, as discussed in Section~\eqref{sub_sec_param}, may result in a duality gap for some cases.  Furthermore, the numerical algorithm we use to find stationary points does not guarantee global optimality.   Despite these issues, the AUC of $97.30\%$ shown in plot (b-3) in Figure~\ref{fig_toy_rev}, indicates a good performance for SBIC.

\begin{figure*}[h]
\centering
		\includegraphics[width=1\textwidth]{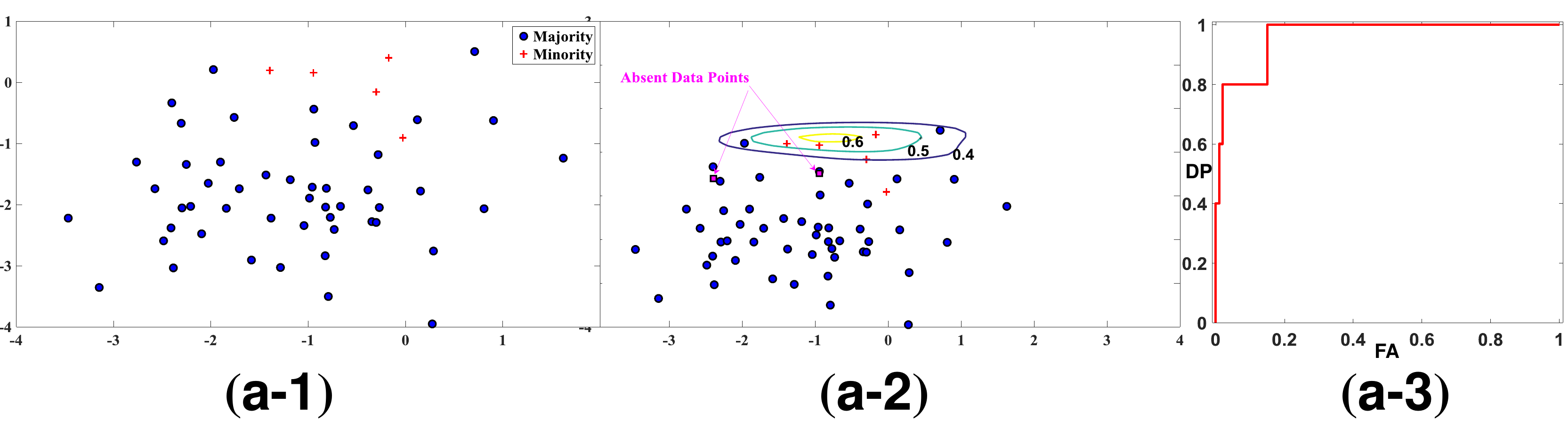}
		\caption {(a-1) Dataset \texttt{Toy3} which suffers from absolute imbalance and overlapping regions. (a-2) The locations of absent data points and the contour plots for the predicted probabilities:  the numbers on the contour curves denote the probability of belonging to the minority class. (a-3) The ROC curve for \texttt{Toy3}.}
\label{fig_toy_rev3}
\end{figure*}

To see how SBIC performs when the datasets are absolutely imbalanced, we create another dataset with the same majority samples and only five minority data points with mean $[0,0]^T$ and the same covariance matrix of the minority as in \texttt{Toy1} and \texttt{Toy2}. \texttt{Toy3} therefore, has the same $70$ majority samples (see plot (a-1) in Figure~\ref{fig_toy_rev3}). We observe that the locations of the absent data points are close to the boundary and away from the minority data point that is within the majority region (plot (a-2) in Figure~\ref{fig_toy_rev3}). This experiment demonstrates the role of absent data points for more challenging cases, i.e. the absent data points try to explore the data region such that they ``push'' the weights towards their optimal values. The AUC of $96.40\%$ shown in plot (a-3) in Figure~\ref{fig_toy_rev3} indicates a good performance for SBIC.  

Next, we examine the effects of parameters $\Delta$ and $\delta$, which appeared in constraints~\eqref{eq:const_min} and ~\eqref{eq:const_maj}, respectively, on the solution of SBIC. Although $\Delta$ and $\delta$ do not appear in equations~\eqref{eq:fonc_x} and~\eqref{eq:fonc_w}, which determine the values of $\mathbf w$ in SBIC, they indirectly impact the solution to~\eqref{eq:fonc_x} and~\eqref{eq:fonc_w} by determining $\lambda_1$ and $\lambda_2$ in optimization problem~\eqref{eq:minR}. Therefore, instead of conducting the sensitivity analysis on the values of $\Delta$ and $\delta$, we conduct it on $\lambda_1$ and $\lambda_2$.

\begin{figure}[h]
\centering
		\includegraphics[width=0.5\textwidth]{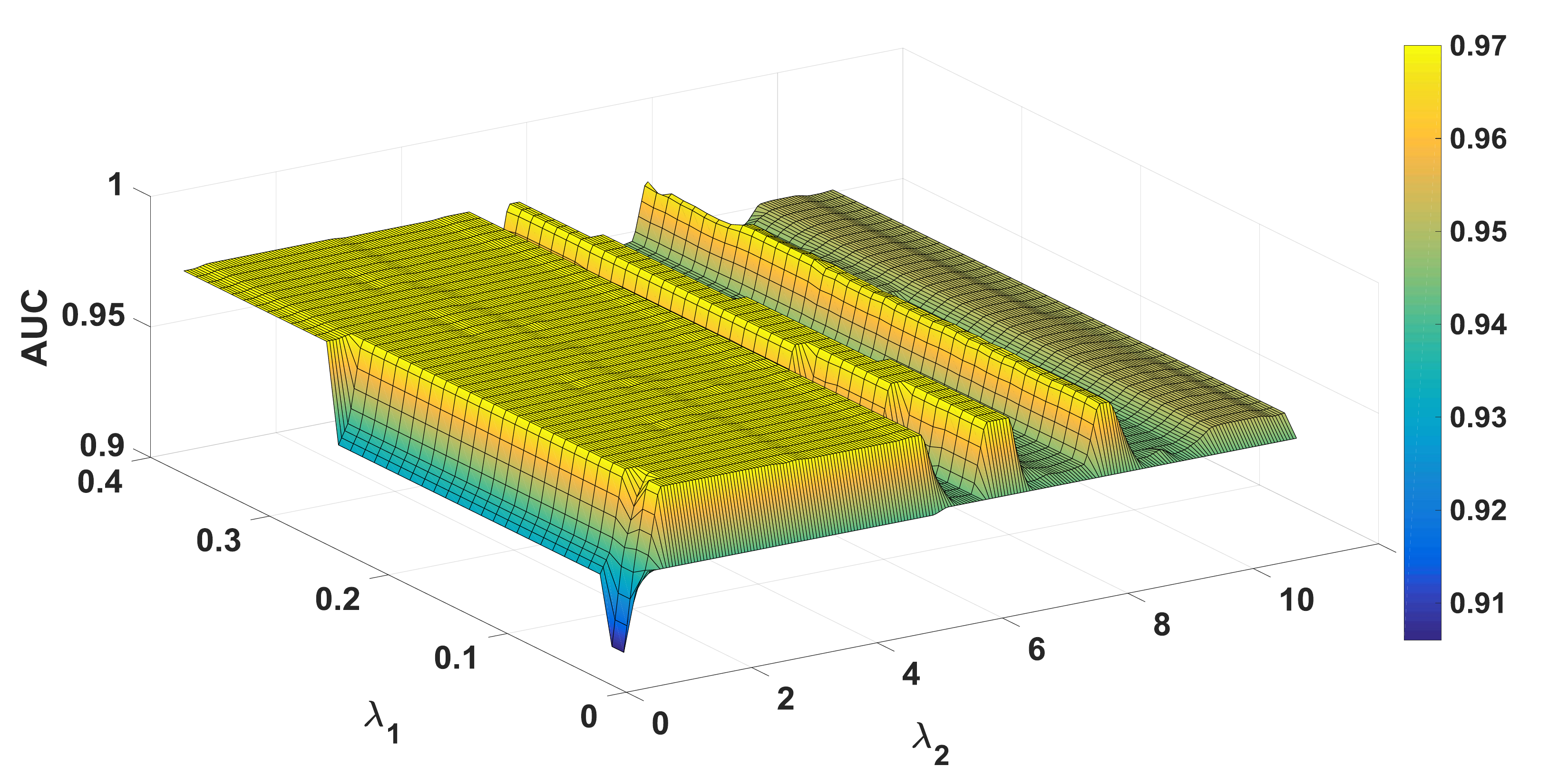}
		\caption {Area  Under Curve using SBIC for different values of $\lambda_1$ and $\lambda_2$ for dataset \texttt{Toy3}.}
\label{fig_toy_rev3_sen}
\end{figure}

Figure~\ref{fig_toy_rev3_sen} shows the AUC for different combinations of $\lambda_1$ and $\lambda_2$ for dataset \texttt{Toy3}. We produce this figure by finding the AUC for a set of $(\lambda_1,\lambda_2)\in[0,4]\times[0,11]$, and then interpolate the results to get a continuous surface for illustration. Figure~\ref{fig_toy_rev3_sen} suggests that when both $\lambda_1$ and $\lambda_2$ are very close to zero, which means we simply perform classification using an empirical similarity function without generating any absent data, SBIC's performance is not good in terms of AUC. A large  increase in $\lambda_1$, while $\lambda_2$ is still close to zero, will have a minor effect on the performance, whereas if $\lambda_1$ is close to zero, increasing $\lambda_2$ will not improve the performance. This contrast demonstrates the relative importance of constraint~\eqref{eq:const_min} over constraint~\eqref{eq:const_maj} in optimization problem~\eqref{eq:log_lik_con}. SBIC performs consistently well for a large range of $\lambda_1$ and a range greater than $0$ and smaller than $6$ for $\lambda_2$, but, its performance deteriorates significantly for some larger values of $\lambda_2$, which is a manifestation of the non-convexity of the objective function in optimization problem~\eqref{eq:minR}. We note that our cross validation technique to find $\lambda_1$ and $\lambda_2$ provides an AUC equal to $96.40\%$, which is very close to the maximum value $97.00\%$, on the plot. The next section discusses the application of the proposed algorithm to real datasets.
 
\subsection{Experimental Results}\label{sub_sec_results}
We apply SBIC to real datasets and compare its performance with Cost-sensitive Support Vector Machine (CSSVM)~\citep{veropoulos1999controlling} and SMOTE~\citep{Chawla:02}. CSSVM is an SVM algorithm designed for imbalanced classification, where the formulation tries a more strict classification for the minority points by assigning a higher penalty to their mis-classification in the training period. SMOTE generates synthetic minority data points by interpolation. We use an SVM algorithm on the balanced dataset (the dataset obtained by adding the synthetic minority data points). Most other algorithms that deal with imbalanced classification can be categorized into cost-sensitive approaches and synthetic data generation. We choose CSSVM to represent the former, and SMOTE+SVM (hereafter, SMOTE) for the latter, maintaining that CSSVM and SMOTE are sufficient for comparing absent data generations with the two major schools of thought.

We use nine real datasets. Five of the datasets, Breast Cancer Detection, Speech Recognition, Yeast, Ionosphere, and Glass are available on the UCI data repository~\citep{Lichman:2013}. The other four, Pima, E-coli, Haberman, and Vehicle are from~\citep{brown:2013}. When a dataset has labels for more than two classes, we randomly select one class as the minority and aggregate the remaining classes as the majority. Since the number of parameters to learn in SBIC is a function of the dimension of the data, and learning them involves solving a nonlinear optimization, we know that SBIC may not obtain a timely optimal solution for some of the higher dimensional datasets.  Thus, for Vehicle and Ionosphere,  we use Principal Component  Analysis (PCA)  for dimensionality reduction~\citep{jolliffe2002principal}.  For the competing algorithms, we always use the original data without dimensionality reduction. Table~\ref{t_prop} summarizes the properties of the datasets.

\begin{table*}[h]\footnotesize
    \caption{Datasets}\label{t_prop}
    \vspace{3 pt}
\centering
    \begin{tabular}{ | l | c  |c | c | c | c |}
    \hline
    Dataset&  original dim.& dim. used in SBIC&\# of data points & \# of maj. & \# of min.   \\ \hline
    
		\texttt{E-coli}  & 9&9 & 336& 301& 35\\ \hline
		\texttt{Ionosphere} & 34&18 & 351& 225& 126 \\ \hline	
		\texttt{Yeast}  & 10&10 & 1484 & 1449& 35\\ \hline
		\texttt{Glass}  & 9&9 & 214 & 197& 17\\ \hline
      \texttt{Speech Recognition}& 10 &10 & 990& 900& 90 \\ \hline            
       \texttt{Haberman}  & 3&3 & 306& 225& 81\\ \hline
			\texttt{Vehicle}& 18&9 & 846 & 634& 212\\ \hline
			\texttt{Breast Cancer} & 9 &9 & 699 & 458& 241 \\ \hline
			\texttt{Pima}  & 8&8 & 768& 500& 268\\ \hline
       
\end{tabular}
\end{table*}

For each dataset we use five-fold cross validation. Therefore, for each algorithm we obtain 5 AUC values for each dataset. Since we perform cross validation, the imbalance ratio, i.e.  the ratio of the number of majority points to the number of minority points, in each training dataset is fixed. Unlike our previous study~\citep{pourhabib2015absent}, we do not create datasets that  are ``absolutely imbalanced'',  i.e., the training  dataset is imbalanced and contains too few data points.  As such, maintaining the same number of minority  data points in each training case results in better AUCs, compared with~\citep{pourhabib2015absent} which has some training cases containing only a few samples of minority points.

Table~\ref{t_param} summarizes the values of parameters used for each dataset.  To find the Lagrangian coefficients, through evaluating $R(\lambda_{1p},\lambda_{2q})$ in equation~\eqref{eq:R},  we use the candidate sets $\Lambda_1=\{0.05,0.1,0.15,0.2,0.25,0.3, 0.35\}$ and $\Lambda_2=\{0.5,1,3,5,7,9, 11\}$, which means $n_c=7$. The larger candidate values chosen for $\lambda_2$ suggest the need to penalize the violation of constraint~\eqref{eq:const_maj} more compared to~\eqref{eq:const_min}, to ensure the absent data points are close to the boundary of the two classes. Simply put, we do not want the existing minority points to lie between absent data and the majority points, but intend to have the absent data lie between the minority and majority points. The choice of the same candidate sets $\Lambda_1$ and $\Lambda_2$ for all the datasets is justified by the fact that we normalize the input data so that $\|x_i\|^2\le 1$ for $i=1,2,\ldots,n$ in all datasets. The values reported in Table~\ref{t_param} are the average values of $\lambda_1$s and $\lambda_2$s for all undersampled datasets $D_{\ell}$, for $\ell=1,\ldots,U$. Hence, some of the values are not among the candidate values in $\Lambda_1$ or $\Lambda_2$. We determine the values of $\Delta$ and $\delta$ according to~\eqref{eq:BDelta} and~\eqref{eq:Sdelta}; refer to Section~\ref{sub_sec_param} for the determination of $K$ and $U$. 

\begin{table*}[h]\footnotesize
    \caption{Parameters in SBIC}\label{t_param}
    \vspace{3 pt}
\centering
    \begin{tabular}{ | l | c  |c | c | c | c | c|}
    \hline
    Dataset&  $\Delta$ & $\delta$& average $\lambda_1$& average $\lambda_2$ & K & $U$   \\ \hline
    
		\texttt{E-coli}  				 &   55.3     &  22.8      &   0.08      &   9.0            &  56      &  8          \\\hline
		\texttt{Ionosphere} 		 &   87.4     &  49.8      &   0.30      &   5.0           &  180      &   1         \\\hline
		\texttt{Yeast}   				 &    3.5    &   2.3     &   0.05      &   3.0           &  56      &     10       \\\hline
		\texttt{Glass} 					 &     6.6   &    2.0    &   0.10      &   9.0           &  50      &      2      \\\hline
\texttt{Speech Recognition}  &    13.2    &    3.6    &   0.10      &   7.0           &  144      &      3      \\\hline      
       \texttt{Haberman}     &    0.4    &    7.6    &   0.10      &   9.0           &  130      &       3     \\\hline
			\texttt{Vehicle}       &    58.1    &     5.5   &   0.08      &   8.3           &  338      &        3    \\\hline
			\texttt{Breast Cancer} &    45.6    &     157.9    &   0.20       &  0.5           &   355     &         2   \\\hline
			\texttt{Pima}          &    18.5    &    29.5    &   0.05      &   11.0           &  400      &         2   \\\hline
       
\end{tabular}
\end{table*}

\begin{figure*}[h!]
\centering
		\includegraphics[width=\textwidth]{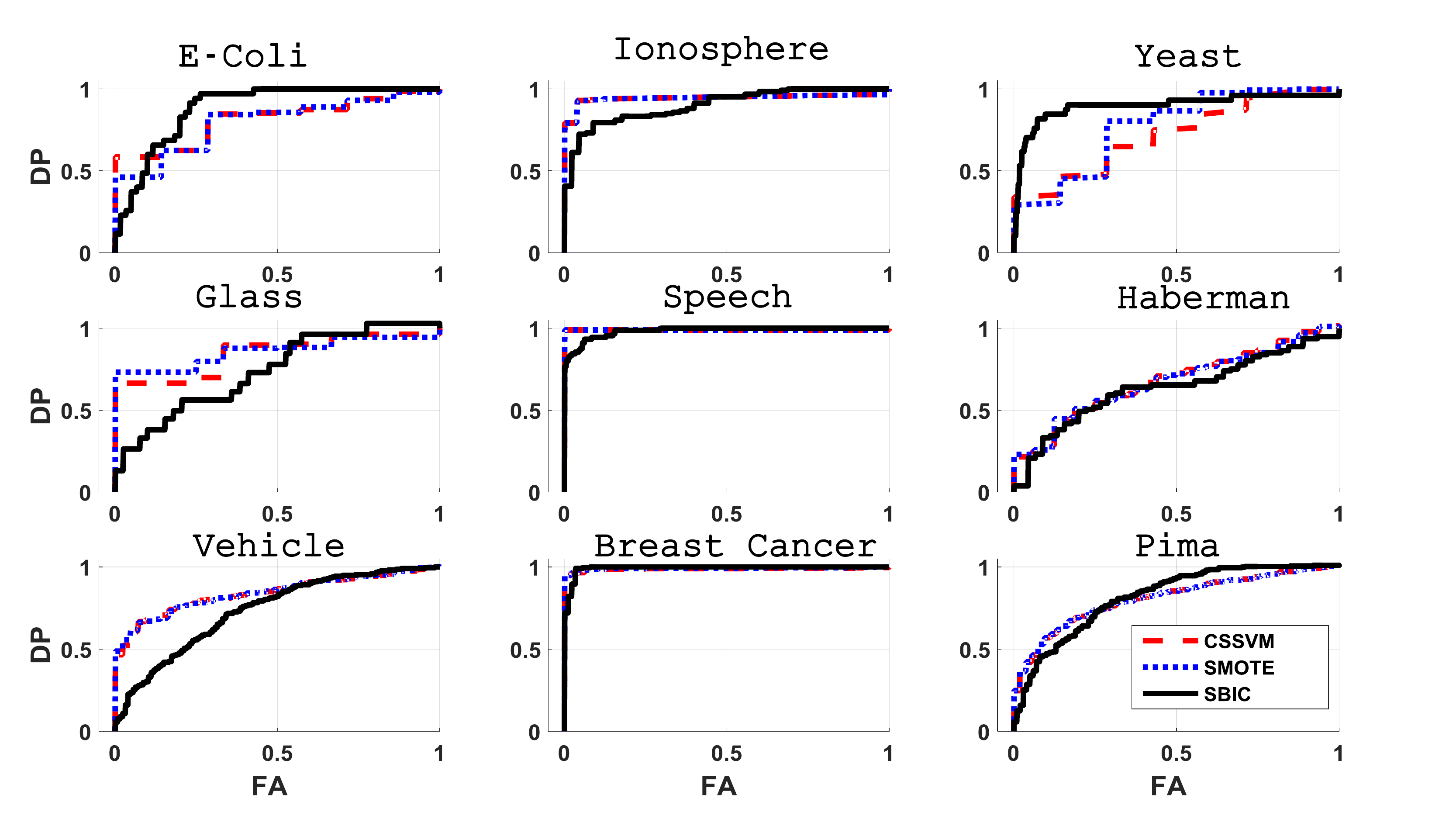}
		\caption {Average ROCs of CSSVM, SMOTE, and SBIC for datasets described in Table~\ref{t_prop}}
\label{fig_ROC}
\end{figure*}

Figure~\ref{fig_ROC} presents the average ROCs of the algorithms for each dataset. We obtain each average ROC by averaging the five curves each associated with one test case (since we do five-fold cross validation). Recall from Section~\ref{sub_sec_eval_crit} that a good way to summarize the information in an ROC curve is to report the area under curve (AUC). Table~\ref{t_res} lists average values of AUC and standard deviations. 

To further illustrate the performance of SBIC, Table~\ref{t_res} presents results for two other algorithms, (1) classification using only an empirical similarity function (ESF)~\citep{gilboa2006empirical} and (2) absent data generation using Fisher discriminant analysis (ADGFDA)~\citep{pourhabib2015absent}. ESF represents an application of the empirical similarity without utilizing any absent data. We include ESF to determine if the inclusion of absent data generation can enhance the performance of a classifier merely based on empirical similarity. We include ADGFDA to compare SBIC with another algorithm that utilizes absent data generation, but inside a different framework, namely Fisher discriminant analysis. 

The results suggest that SBIC outperforms both CSSVM and SMOTE for E-coli, Yeast, Breast Cancer Detection, and Pima. SBIC is competitive with CSSVM and SMOTE for Speech Recognition, but it performs poorly for Ionosphere, Vehicle, and Haberman. ESF also performs poorly on most datasets, unless the dataset is not highly imbalanced (Breast Cancer Detection) or if the classes are well separated (Speech Recognition). We conclude that SBIC’s performance can be attributed mostly to absent data generation rather than to the use of an empirical similarity function

Comparing ADGFDA  with SBIC, on the other hand, does not provide a straightforward conclusion. For some datasets (Yeast, E-coli), ADGFDA  and SBIC outperform the competing algorithms, whereas for Ionosphere and Vehicle, ADGFDA  and SBIC do not outperform CSSVM and SMOTE. The results suggest that for the former subset of datasets, absent data generation can improve the mis-classification rate and for the latter, absent data generation can be inadequate (as opposed to cost-sensitive or synthetic data generation). In other words, absent data generation may not help improve a classifier performance for some unbalanced data structures.

Another group of datasets, (Pima, Glass) show a discrepancy between the performance of ADGFDA  and SBIC. We explain the discrepancy due to the different base classifiers, namely Fisher discriminate in ADGFDA and empirical similarity in SBIC. The results suggest that Fisher discriminant analysis may be better suited to some data structures compared to an empirical similarity function.

Note that the average ROCs are obtained by averaging the curves (which involves interpolation and therefore approximation), whereas the average AUCs reported in Table~\ref{t_res} are obtained by averaging the AUCs under the five curves for each dataset. As such there might be a slight difference between the actual AUC shown  in Figure~\ref{fig_ROC} and the average AUCs reported in Table~\ref{t_res}. For most cases however, the difference is insignificant.
 
\begin{table*}\footnotesize
    \caption{Average AUCs. The numbers in parentheses are standard deviations of five folds.}\label{t_res}
    \vspace{3 pt}
\centering
    \begin{tabular}{ | l | c | c | c | c| c|}
    \hline
    Dataset&  CSSVM& SMOTE&ESF &   ADGFDA       &        SBIC 	 \\ \hline	
	\texttt{E-coli}  & 81.58  (8.7)  & 79.82  (8.2)  & 76.39  (8.3)  & 83.11  (9.1)  & {\bf 88.59  (5.5)}  \\ \hline	
\texttt{Ionosphere} &{\bf 94.24  (2.0)}  & {\bf 94.24  (2.1)}  & 84.38  (5.54)  & 90.5   (1.4)  & 90.85  (7.3)  \\\hline	
\texttt{Yeast}  & 72.05  (17.3) & 77.05  (12.9) & 79.38  (5.5)  & 89.42  (11.7) & {\bf 89.76  (4.7)}  \\\hline	
\texttt{Glass}  & 84.63  (19.2) & 85.81  (11.1) & 69.5   (10.9) & {\bf 87.82  (13.6)} & 74.4   (11.7) \\ \hline	
\texttt{Speech Recognition}& 98.86  (0.7)  & 98.66  (0.4)  & 98.79  (0.01)  & {\bf 99.08  (0.87)}  & 98.38  (1.11)  \\\hline	
\texttt{Haberman} &  68.03  (3.9)  & 67.62  (3.1)  & 60.76  (7.9)  & {\bf 69.23  (7.6)}  & 63.79  (7.1)  \\\hline	
\texttt{Vehicle}& {\bf 84.57  (4.3)}  & 84.38  (4.1)  & 67.71  (5.9)  & 79.49  (4.2)  & 73.73  (6.7)   \\\hline	
\texttt{Breast Cancer} & 98.86  (0.7)  & 99.06  (1.02)  & 98.74  (0.5)  & {\bf 99.33}  (0.7)  & 99.29  (0.5)  \\\hline	
\texttt{Pima}  & 81.42  (1.4)  & 81.42  (1.4)  & 75.18  (3.1)  & 74.01  (5.3)  & {\bf 82.39  (2.5)}\\\hline	
\end{tabular}
\end{table*}

\begin{figure}[h]
\centering
\includegraphics[width=0.5\textwidth]{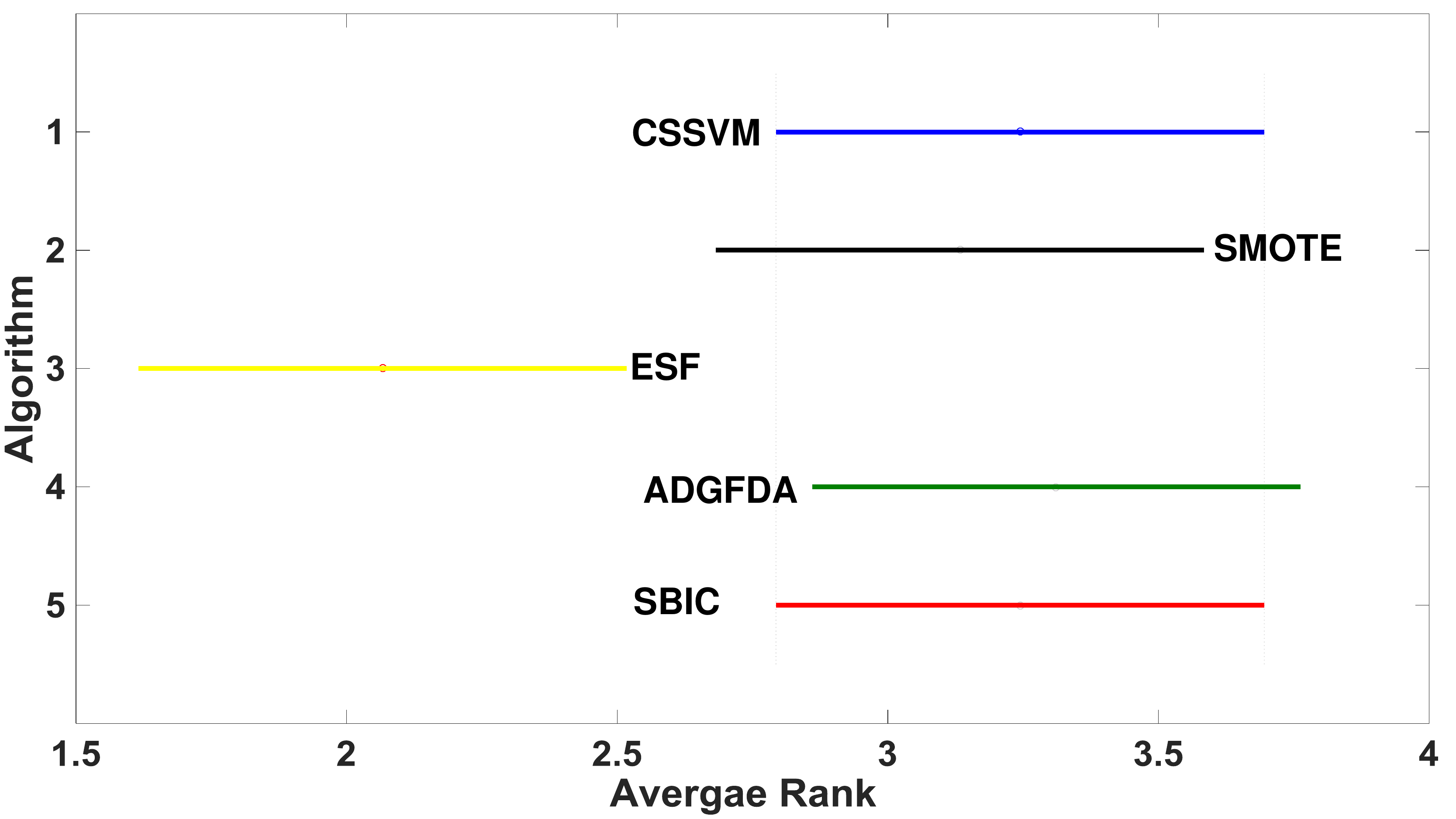}
\caption{Posthoc analysis on the ranking data. The bars denote approximately $95\%$ confidence intervals.}
\label{fig_friedman}
\end{figure}

Acknowledging that SBIC outperforms the competing methods for some datasets, now we need to determine if the results are statistically significant based on the nine datasets. Using the data reported in Table 3, we conduct a posthoc analysis using the Friedman test~\citep{demsar_statistical_2006} to rank the algorithms. We let $m_d$ denote the number of test sets and $m_a$ denote the number of algorithms. We define $\mathbf R$  as an $m_d\times m_a$ matrix whose $(i,j)$th entry denotes the AUC of algorithm $j$ for the test set $i$, where $j=1,\ldots,m_a$ and $i=1,\ldots,m_d$. Based on the data in matrix $\mathbf R$ we create another matrix $\mathbf Q$ of the same size whose $(i,j)$th entry denotes the rank for the algorithm $j$ for the test set $i$, i.e., each row in the matrix $\mathbf Q$ denotes the rank of each algorithm for that test set, where the best algorithm has rank $m_a$ and the worst has rank $1$. We let $\overline{\mathbf q}$ denote an $m_a\times 1$ vector whose $\ell$th entry $\overline{\mathbf q}(\ell)$ is the average value of the $\ell$th column of $\mathbf Q$. Under the null hypothesis that all algorithms are equivalent and in the sense that for a given dataset they produce the same AUC, the Friedman statistic
\begin{equation}
\mathcal{F}=\frac{12m_{d}}{m_{a}(m_{a}+1)}\left(\sum_{\ell=1}^{m_{a}}\overline{\mathbf q}(\ell)^2-\frac{m_{a}(m_{a}+1)^2}{4}\right),
\end{equation}
follows a Chi-squared distribution with $m_{a}-1$ degrees of freedom. Here, we have $m_a=5$ algorithms and nine datasets, but since we do a five-fold cross validation, we have $m_d=9\times 5=45$ test sets. Therefore, matrix $\mathbf R$ is a $45\times 5$. That is, $\mathbf R$ is the expanded form of the results in Table~\ref{t_res} where each row in the tables is expanded into five rows for the matrix $\mathbf R$. Table~\ref{tab_friedman} presents the average rankings based on the Friedman test, where  $5$ is the ranking of the best algorithm. Figure~\ref{fig_friedman} displays the posthoc analysis on the results of the test. According to Table~\ref{tab_friedman}, the average ranking for ADGFDA is the highest among the five algorithms, and SBIC and CSSVM are tied for second place. Figure~\ref{fig_friedman} shows that the difference between CSSVM, SMOTE, ADGFDA, and SBIC is not statistically significant (based on the nine datasets used and five-fold cross validation). In fact, the only significant result is that ESF performs the worst.

\begin{table}\footnotesize
\centering
    \begin{tabular}{ | l | c | c | c | c | c| }
    \hline
    Algorithm& CSSVM& SMOTE& ESF & ADGFDA  & SBIC\\\hline
    Ranking Mean& 3.24 &3.13 &2.07&{\bf 3.31}&3.24  \\ \hline
\end{tabular}
    \caption{Ranking mean for the algorithms based on Friedman test}\label{tab_friedman}
\end{table}

In summary, despite not being the statistically superior algorithm in this study, SBIC does outperform competing algorithms, in some cases quite remarkably, on some of the datasets reported in Table~\ref{t_res}. This shows SBIC has some merits for imbalanced classification, and can be considered as a viable alternative, at least for some data structures, to traditional state-of-the-art algorithms such as CSSVM.

\section{Conclusion}\label{sec_concl}

Imbalanced classification is of paramount importance in applications such as quality control, healthcare informatics, and warranty claims. This paper has proposed an absent data generation  mechanism based on empirical similarity  for imbalanced classification. The approach falls in the category of synthetic data generation mechanisms that are embedded in the classification algorithms, namely absent data generation. The proposed algorithm, SBIC, does not actually generate synthetic data, but instead utilizes their properties to identify the weights of an empirical similarity function.

We formulated the imbalance classification problem as a constrained optimization framework and used numerical techniques to find the solution.  Based on empirical studies of nine real datasets, we found that SBIC outperformed the other commonly used algorithms for some datasets. A failure to outperform was attributed to the fact that absent data generation  does not necessarily improve a classifying algorithm's prediction power, or to the specific mechanism for absent data generation employed in SBIC. SBIC was also limited by the ``manual'' selection of some parameters,  such as $\delta$, $\Delta$, or $T$, which suggested that an automated approach for selecting parameters could potentially improve algorithmic performance.

The limitations above suggest four paths for future research on SBIC. First, the imbalanced classification literature would benefit from a thorough study that determines the applicability  of synthetic data generation, in general, and absent  data generation, in particular, to imbalanced datasets. Our review of the published literature found that studies focus primarily  on empirical results, whereas establishing a theoretical foundation that connects the data structure to the algorithms would provide insights into improving the design of the SBIC algorithm for imbalanced classification. Second, SBIC should be tested on more absolutely imbalanced datasets for which we have only a few samples from the minority class, by either exploring other datasets or creating training datasets through undersampling~\citep{pourhabib2015absent}. Third,  the application of variable-bandwidth kernels~\citep{giannakis2012nonlinear} to imbalanced classification may proved useful for imbalance classification because the kernels tend to be more stable in regions with low sample density.  Fourth,  since the specific structure of spatio-temporal data hinders a direct application of absent  data generation techniques, it would be worthwhile to determine the applicability  of imbalanced classification techniques to spatio-temporal datasets. From a data mining perspective, however, rare-events in spatio-temporal systems~\citep{giannakis2012nonlinear} can be categorized as minority  data points.  Extending similarity-based absent  data generation to such problems, while not straightforward should be an ongoing pursuit.

  \section*{Acknowledgment}
The research was partly supported  by OSU Foundation for the National Energy Solutions Institute - Smart Energy Source, grant 20-96680 . This work was completed utilizing the High Performance Computing Center facilities of Oklahoma State University at Stillwater.




%
\bibliographystyle{chicago}
\bibliography{IEEEKDE_bib}

\begin{thebibliography}{}

\bibitem[\protect\citeauthoryear{??}{bro}{2014}]{brown:2013}
 (2014).
\newblock Center for evidence-based medicine.
\newblock \url{http://www.cebm.brown.edu/static/imbalanced-datasets.zip }.
\newblock last accessed 07/2014.

\bibitem[\protect\citeauthoryear{Barua, Islam, Yao, and Murase}{Barua
  et~al.}{2014}]{barua2014mwmote}
Barua, S., M.~M. Islam, X.~Yao, and K.~Murase (2014).
\newblock {MWMOTE}--majority weighted minority oversampling technique for
  imbalanced data set learning.
\newblock {\em IEEE Transactions on Knowledge and Data Engineering\/}~{\em
  26\/}(2), 405--425.

\bibitem[\protect\citeauthoryear{Billot, Gilboa, and Schmeidler}{Billot
  et~al.}{2008}]{billot2008axiomatization}
Billot, A., I.~Gilboa, and D.~Schmeidler (2008).
\newblock Axiomatization of an exponential similarity function.
\newblock {\em Mathematical Social Sciences\/}~{\em 55\/}(2), 107--115.

\bibitem[\protect\citeauthoryear{Bradley}{Bradley}{1997}]{bradley1997use}
Bradley, A.~P. (1997).
\newblock The use of the area under the {ROC} curve in the evaluation of
  machine learning algorithms.
\newblock {\em Pattern Recognition\/}~{\em 30\/}(7), 1145--1159.

\bibitem[\protect\citeauthoryear{Byon, Shrivastava, and Ding}{Byon
  et~al.}{2010}]{byon:10}
Byon, E., A.~K. Shrivastava, and Y.~Ding (2010).
\newblock A classification procedure for highly imbalanced class sizes.
\newblock {\em IIE Transactions\/}~{\em 42\/}(4), 288--303.

\bibitem[\protect\citeauthoryear{Byrd, Gilbert, and Nocedal}{Byrd
  et~al.}{2000}]{byrd2000trust}
Byrd, R.~H., J.~C. Gilbert, and J.~Nocedal (2000).
\newblock A trust region method based on interior point techniques for
  nonlinear programming.
\newblock {\em Mathematical Programming\/}~{\em 89\/}(1), 149--185.

\bibitem[\protect\citeauthoryear{Chawla, Bowyer, Hall, and Kegelmeyer}{Chawla
  et~al.}{2002}]{Chawla:02}
Chawla, N.~V., K.~W. Bowyer, L.~O. Hall, and W.~P. Kegelmeyer (2002).
\newblock {SMOTE}: {S}ynthetic {M}inority {O}ver-sampling {T}echnique.
\newblock {\em Journal of Artificial Intelligence Research\/}~{\em 16},
  321--357.

\bibitem[\protect\citeauthoryear{Chen, Tsai, Young, and Kodell}{Chen
  et~al.}{2005}]{chen:05}
Chen, J.~J., C.~A. Tsai, J.~F. Young, and R.~L. Kodell (2005).
\newblock Classification ensembles for unbalanced class sizes in predictive
  toxicology.
\newblock {\em SAR and QSAR in Environmental Research\/}~{\em 16\/}(6),
  517--529.

\bibitem[\protect\citeauthoryear{Chen, He, and Garcia}{Chen
  et~al.}{2010}]{chen2010ramoboost}
Chen, S., H.~He, and E.~A. Garcia (2010).
\newblock Ramoboost: Ranked minority oversampling in boosting.
\newblock {\em IEEE Transactions on Neural Networks\/}~{\em 21\/}(10),
  1624--1642.

\bibitem[\protect\citeauthoryear{Conn, Gould, and Toint}{Conn
  et~al.}{2000}]{conn2000trust}
Conn, A.~R., N.~I.~M. Gould, and P.~L. Toint (2000).
\newblock {\em Trust-Region Methods}.
\newblock SIAM.

\bibitem[\protect\citeauthoryear{de~Mantaras and Armengol}{de~Mantaras and
  Armengol}{1998}]{de1998machine}
de~Mantaras, R.~L. and E.~Armengol (1998).
\newblock Machine learning from examples: Inductive and lazy methods.
\newblock {\em Data \& Knowledge Engineering\/}~{\em 25\/}(1), 99--123.

\bibitem[\protect\citeauthoryear{Dem\u{s}ar}{Dem\u{s}ar}{2006}]{demsar_statistical_2006}
Dem\u{s}ar, J. (2006).
\newblock Statistical comparisons of classifiers over multiple data sets.
\newblock {\em The Journal of Machine Learning Research\/}~{\em 7}, 1–--30.

\bibitem[\protect\citeauthoryear{Efron}{Efron}{1982}]{efron1982jackknife}
Efron, B. (1982).
\newblock The jackknife, the bootstrap and other resampling plans.
\newblock In {\em CBMS-NSF Regional Conference Series in Applied Mathematics},
  Volume~38. SIAM.

\bibitem[\protect\citeauthoryear{Elkan}{Elkan}{2001}]{Elkan:01}
Elkan, C. (2001).
\newblock The foundations of cost-sensitive learning.
\newblock In {\em Proceedings of the Seventeenth International Joint Conference
  on Artificial Intelligence}, pp.\  973--978.

\bibitem[\protect\citeauthoryear{Galar, Fernandez, Barrenechea, Bustince, and
  Herrera}{Galar et~al.}{2012}]{galar2012review}
Galar, M., A.~Fernandez, E.~Barrenechea, H.~Bustince, and F.~Herrera (2012).
\newblock A review on ensembles for the class imbalance problem: bagging-,
  boosting-, and hybrid-based approaches.
\newblock {\em IEEE Transactions on Systems, Man, and Cybernetics, Part C:
  Applications and Reviews\/}~{\em 42\/}(4), 463--484.

\bibitem[\protect\citeauthoryear{Giannakis and Majda}{Giannakis and
  Majda}{2012}]{giannakis2012nonlinear}
Giannakis, D. and A.~J. Majda (2012).
\newblock Nonlinear laplacian spectral analysis for time series with
  intermittency and low-frequency variability.
\newblock {\em Proceedings of the National Academy of Sciences\/}~{\em
  109\/}(7), 2222--2227.

\bibitem[\protect\citeauthoryear{Gilboa, Lieberman, and Schmeidler}{Gilboa
  et~al.}{2006}]{gilboa2006empirical}
Gilboa, I., O.~Lieberman, and D.~Schmeidler (2006).
\newblock Empirical similarity.
\newblock {\em The Review of Economics and Statistics\/}~{\em 88\/}(3),
  433--444.

\bibitem[\protect\citeauthoryear{Gilboa, Lieberman, and Schmeidler}{Gilboa
  et~al.}{2011}]{gilboa2011similarity}
Gilboa, I., O.~Lieberman, and D.~Schmeidler (2011).
\newblock A similarity-based approach to prediction.
\newblock {\em Journal of Econometrics\/}~{\em 162\/}(1), 124--131.

\bibitem[\protect\citeauthoryear{Han, Wang, and Mao}{Han et~al.}{2005}]{Han:05}
Han, H., W.-Y. Wang, and B.-H. Mao (2005).
\newblock Borderline-{SMOTE}: A new over-sampling method in imbalanced data
  sets learning.
\newblock In {\em Advances in Intelligent Computing}, Volume 3644 of {\em
  Lecture Notes in Computer Science}, pp.\  878--887. Springer Berlin
  Heidelberg.

\bibitem[\protect\citeauthoryear{Hastie, Tibshirani, and Friedman}{Hastie
  et~al.}{2009}]{hastie:09}
Hastie, T., R.~Tibshirani, and J.~Friedman (2009).
\newblock {\em The Elements of Statistical Learning: Data Mining, Inference,
  and Prediction, Second Edition}.
\newblock Springer Series in Statistics. New York, NY, USA: Springer New York
  Inc.

\bibitem[\protect\citeauthoryear{He and Garcia}{He and Garcia}{2009}]{He:09}
He, H. and E.~A. Garcia (2009).
\newblock Learning from imbalanced data.
\newblock {\em IEEE Transactions on Knowledge and Data Engineering\/}~{\em
  21\/}(9), 1263--1284.

\bibitem[\protect\citeauthoryear{Jolliffe}{Jolliffe}{2002}]{jolliffe2002principal}
Jolliffe, I. (2002).
\newblock {\em Principal Component Analysis (second edition)}.
\newblock New York: Springer.

\bibitem[\protect\citeauthoryear{Kanungo, Mount, Netanyahu, Piatko, Silverman,
  and Wu}{Kanungo et~al.}{2002}]{kanungo2002efficient}
Kanungo, T., D.~M. Mount, N.~S. Netanyahu, C.~D. Piatko, R.~Silverman, and
  A.~Y. Wu (2002).
\newblock An efficient k-means clustering algorithm: Analysis and
  implementation.
\newblock {\em IEEE Transactions on Pattern Analysis and Machine
  Intelligence\/}~{\em 24\/}(7), 881--892.

\bibitem[\protect\citeauthoryear{Lichman}{Lichman}{2013}]{Lichman:2013}
Lichman, M. (2013).
\newblock {UCI} machine learning repository.

\bibitem[\protect\citeauthoryear{Liu, Wu, and Zhou}{Liu
  et~al.}{2009}]{liu2009exploratory}
Liu, X.-Y., J.~Wu, and Z.-H. Zhou (2009).
\newblock Exploratory undersampling for class-imbalance learning.
\newblock {\em IEEE Transactions on Systems, Man, and Cybernetics, Part B:
  Cybernetics\/}~{\em 39\/}(2), 539--550.

\bibitem[\protect\citeauthoryear{Masnadi-Shirazi and
  Vasconcelos}{Masnadi-Shirazi and Vasconcelos}{2010}]{masnadi2010risk}
Masnadi-Shirazi, H. and N.~Vasconcelos (2010).
\newblock Risk minimization, probability elicitation, and cost-sensitive svms.
\newblock In {\em Proceedings of the 27th International Conference on Machine
  Learning (ICML-10)}, pp.\  759--766.

\bibitem[\protect\citeauthoryear{Mika, R\"{a}tsch, Weston, Sch\"{o}lkopf, and
  M\"{u}llers}{Mika et~al.}{1999}]{Mika:99}
Mika, S., G.~R\"{a}tsch, J.~Weston, B.~Sch\"{o}lkopf, and K.-R. M\"{u}llers
  (1999, Aug).
\newblock Fisher discriminant analysis with kernels.
\newblock In {\em Neural Networks for Signal Processing IX, 1999. Proceedings
  of the 1999 IEEE Signal Processing Society Workshop.}, pp.\  41 --48.

\bibitem[\protect\citeauthoryear{Park, Huang, and Ding}{Park
  et~al.}{2010}]{Park:10}
Park, C., J.~Z. Huang, and Y.~Ding (2010, Sep).
\newblock A computable plug-in estimator of minimum volume sets for novelty
  detection.
\newblock {\em Operations Research\/}~{\em 58\/}(5), 1469--1480.

\bibitem[\protect\citeauthoryear{Pourhabib, Mallick, and Ding}{Pourhabib
  et~al.}{2015}]{pourhabib2015absent}
Pourhabib, A., B.~K. Mallick, and Y.~Ding (2015).
\newblock Absent data generating classifier for imbalanced class sizes.
\newblock {\em The Journal of Machine Learning Research\/}~{\em 16},
  2695--2724.

\bibitem[\protect\citeauthoryear{Ramentol, Caballero, Bello, and
  Herrera}{Ramentol et~al.}{2012}]{ramentol2012smote}
Ramentol, E., Y.~Caballero, R.~Bello, and F.~Herrera (2012).
\newblock {SMOTE-RSB*}: a hybrid preprocessing approach based on oversampling
  and undersampling for high imbalanced data-sets using smote and rough sets
  theory.
\newblock {\em Knowledge and Information Systems\/}~{\em 33\/}(2), 245--265.

\bibitem[\protect\citeauthoryear{Sun, Kamel, Wong, and Wang}{Sun
  et~al.}{2007}]{sun2007cost}
Sun, Y., M.~S. Kamel, A.~K. Wong, and Y.~Wang (2007).
\newblock Cost-sensitive boosting for classification of imbalanced data.
\newblock {\em Pattern Recognition\/}~{\em 40\/}(12), 3358--3378.

\bibitem[\protect\citeauthoryear{Ting}{Ting}{2002}]{Ting:02}
Ting, K.~M. (2002).
\newblock An instance-weighting method to induce cost-sensitive trees.
\newblock {\em IEEE Transactions on Knowledge and Data Engineering\/}~{\em
  14\/}(3), 659--665.

\bibitem[\protect\citeauthoryear{Veropoulos, Campbell, Cristianini,
  et~al.}{Veropoulos et~al.}{1999}]{veropoulos1999controlling}
Veropoulos, K., C.~Campbell, N.~Cristianini, et~al. (1999).
\newblock Controlling the sensitivity of support vector machines.
\newblock In {\em Proceedings of the International Joint Conference on AI},
  pp.\  55--60.

\bibitem[\protect\citeauthoryear{Zhou and Liu}{Zhou and
  Liu}{2006}]{zhou2006training}
Zhou, Z.-H. and X.-Y. Liu (2006).
\newblock Training cost-sensitive neural networks with methods addressing the
  class imbalance problem.
\newblock {\em IEEE Transactions on Knowledge and Data Engineering\/}~{\em
  18\/}(1), 63--77.

\end{thebibliography}

%




\end{document}